\documentclass{article}

\usepackage[final]{corl_2022} %

\title{PRISM: Probabilistic Real-Time Inference in \\ Spatial World Models}

\author{
    Atanas Mirchev{\normalfont\textsuperscript{1},} Baris Kayalibay{\normalfont\textsuperscript{1},} Ahmed Agha{\normalfont\textsuperscript{1},}\\
    \textbf{Patrick van der Smagt}\textsuperscript{1}, \textbf{Daniel Cremers}\textsuperscript{2}, \textbf{Justin Bayer}\textsuperscript{1}\\
    \\
    \textsuperscript{1}Machine Learning Research Lab, Volkswagen Group, \textsuperscript{2}Technical University of Munich\\
    \texttt{atanas.mirchev@argmax.ai}
}

\usepackage[utf8]{inputenc} %
\usepackage[T1]{fontenc}    %
\usepackage{url}            %
\usepackage{booktabs}       %
\usepackage{amsfonts}       %
\usepackage{nicefrac}       %
\usepackage{microtype}      %
\usepackage[dvipsnames]{xcolor}         %

\usepackage{amsmath}
\usepackage{todonotes}
\usepackage{mymath}
\usepackage{cleveref}
\usepackage{subcaption}
\usepackage{caption}
\usepackage{tikz}
\usepackage{tabularx}
\usepackage{multicol}
\usepackage{multirow}
\usepackage{booktabs}
\usepackage{mlmacros}
\usepackage{tablefootnote}
\usepackage{scrextend}
\usepackage{siunitx}
\usepackage{enumitem}
\usepackage{wrapfig}
\usepackage{float}

\hypersetup{
    colorlinks,
    linkcolor={blue!50!black},
    citecolor={blue!50!black},
    urlcolor={blue!50!black},
}
\usetikzlibrary{positioning,shapes,arrows}

\definecolor{myblue}{RGB}{52, 152, 219}
\def\cf{c.f.\,}
\def\ie{i.e.\,}
\def\eg{e.g.\,}
\def\wrt{w.r.t.\,}
\def\methodname{PRISM}

\begin{document}
\maketitle

\begin{abstract}
    We introduce \methodname, a method for real-time filtering in a probabilistic generative model of agent motion and visual perception.
    Previous approaches either lack uncertainty estimates for the map and agent state, do not run in real-time, do not have a dense scene representation or do not model agent dynamics.
    Our solution reconciles all of these aspects.
    We start from a predefined state-space model which combines differentiable rendering and 6-DoF dynamics.
    Probabilistic inference in this model amounts to simultaneous localisation and mapping (SLAM) and is intractable.
    We use a series of approximations to Bayesian inference to arrive at probabilistic map and state estimates.
    We take advantage of well-established methods and closed-form updates, preserving accuracy and enabling real-time capability.
    The proposed solution runs at 10Hz real-time and is similarly accurate to state-of-the-art SLAM in small to medium-sized indoor environments, with high-speed UAV and handheld camera agents (Blackbird, EuRoC and TUM-RGBD).
\end{abstract}
\section{Introduction} \label{sec:introduction}
Moving agents perceive streams of information, typically a mix of RGB images, depth and inertial measurements.
Probabilistic generative models \citep{koller2009probabilistic} are a principled way to formalise the \emph{synthesis} of this data, and from these models inference can be derived through Bayes' rule.
We focus on exactly such inference and target the agent states and the scene map, a problem known as simultaneous localisation and mapping (SLAM).
We treat it as a posterior approximation for a given state-space model, such that the combination is useful for model-based control: the posterior inference serves as a state estimator and the predictive state-space model as a simulator with which to plan ahead \citep{bertsekas}.

To pave the way towards decision making, we believe an inference method should have:
\begin{itemize}[topsep=0pt]
    \itemsep0em
    \item a compatible predictive model for both RGB-D images and 6-DoF dynamics;
    \item principled state and map uncertainty;
    \item real-time performance on commodity hardware;
    \item state-of-the-art localisation accuracy.
\end{itemize}
We motivate these requirements further in \cref{app:motivation}.
Prominent methods like LSD-SLAM \citep{engel2014lsd}, ORB-SLAM \citep{orbslam2}, DSO \citep{dso} have propelled visual SLAM forward, with heavy focus on large-scale localisation. 
The core of modern large-scale SLAM is maximum a-posteriori (MAP) smoothing in a probabilistic factor graph \cite{cadena2016past,kschischang2001factor}.
At present this demands sparsity assumptions for computational feasibility, which obstructs the tight integration of dense maps and rendering.
Nonetheless, for smaller scenes the recent popularity of neural models (\eg NERF~\citep{nerf}) has sparked interest in inference through a renderer (e.g.\ \citep{Zhu2022CVPR,koestler2021tandem,sucar2021imap}), but dynamics modelling and uncertainty have remained out of scope.
Conversely, classical filtering comes with dynamics and uncertainty in real-time (e.g.\ \citep{kalman1960new,fastslam,grisetti2007improved}), but over time has given way to large-scale smoothing \citep{cadena2016past} and to our knowledge has not been well explored for the integration of dense differentiable rendering and dynamics on a moderate scale.

\begin{figure}[t]
    \centering
    \includegraphics[width=0.7\textwidth]{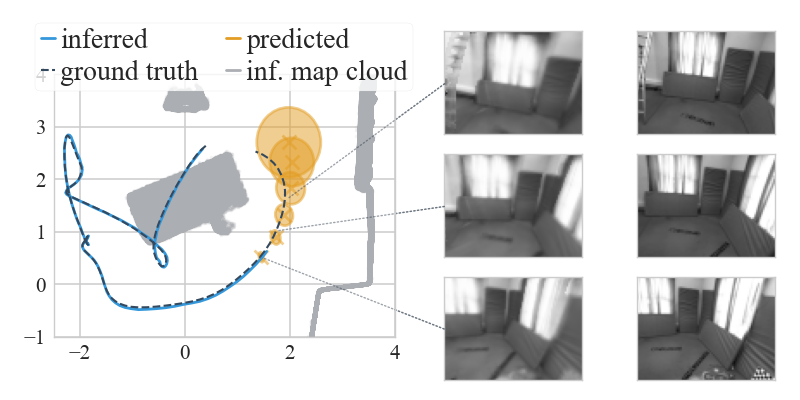}
    \caption{
        Inference is tailored to the depicted predictive model.
        Predicting future rollouts, as shown, is required for optimal control.
        Ground-truth trajectory in \emph{black}, inferred trajectory from past data in \color{myblue}\emph{blue}\color{black}.
        In \color{orange}\emph{orange}\color{black}, we see uncertainty envelopes for the predicted future states.
        On the right, we see predicted and ground-truth future images.
        Visualised in 2D for clarity, our method operates in 3D.
    }
    \label{fig:predictive}
\end{figure}

Overall, we find there is a need for a cohesive inference solution that satisfies our requirements.
We thus contribute by meeting all the above goals, emphasising the link to a predictive model (\cref{fig:predictive}).

We start from the generative model of \citet{mirchev2021variational}, who combine differentiable rendering and agent dynamics in a probabilistic framework.
The authors considered stochastic variational inference for this model, applying it off-line with runtime orders of magnitude too long for on-line use.
We pursue an alternative route for real-time inference: from the generative assumptions we derive approximations to the true marginal filters over the last state and map \citep{sarkka2013bayesian}.
By focusing on recursive filtering updates, we identify where established probabilistic inference and computer vision techniques can be used, putting emphasis on fast closed-form updates.
We find this divide-and-conquer strategy is a good compromise for achieving the aforementioned objectives under computational constraints.

We evaluate the proposed solution on two unmanned aerial vehicle (UAV) data sets \citep{euroc,blackbird} and on TUM-RGBD \citep{tumrgbd}.
Our method PRISM runs at 10\,Hz real-time with similar localisation accuracy to state-of-the-art SLAM in moderately-sized indoor environments.
It provides uncertainty estimates and features a predictive distribution that can both render images and forecast the agent's movement.
\section{Related Work}

\paragraph{Generative models} Generative state-space models simulate the formation of observed data over time in a Markov chain \citep{koller2009probabilistic,kalman1960new,hmms,dvbf,dkf,vrnn}, serving as \emph{world} models \citep{envsims,worldmodels}.
With their agent dynamics and state-to-observation emission models we can imagine future rollouts for planning \citep{bertsekas,pilco,planet,dreamer,slac,learntofly,empowerment,vast,deepmbrl}.
We abide by this framework and design a posterior inference for a \emph{spatial} state-space model, to enable on-line control.
Among such models (e.g. \citep{gtmsm,gqn,gregor2019,abise,gupta2019cognitive,parisotto2018neural,chaplot2020learning}), we tailor our inference to the model of \citet{mirchev2021variational}.
It scales to 3D with rendering and 6-DoF dynamics.
We contribute a real-time inference that fits its probabilistic formulation.
\paragraph{SLAM through image synthesis}
The assumed generative model renders RGB-D images, which is related to SLAM through full-image synthesis.
Traditional methods feature varied maps, from volumetric to surfels (e.g. \citep{curless1996volumetric,carr2001reconstruction,dtam,newcombe2011kinectfusion,niessner2013real,keller2013realtime,Whelan2015ElasticFusionDS,Cao2018RealtimeHT,hrbf}), and commonly estimate new camera poses by aligning new observations to a rendered image with variants of point-to-plane ICP with photometric consistency \citep{chen1992object,steinbrucker2011real,audras2011real,kerl2013dense,newcombe2011kinectfusion}.
We extend this optimisation with dynamics in our approximate state filter \citep{kayalibay2022tracking}.
A recent trend is to use implicit scene representations like NERF (e.g. \citep{nerf,siren,isdf}) with high rendering fidelity.
Gradient-based pose inference through NERF-like rendering has received attention \citep{inerf,nerfmm}, with iMAP \citep{sucar2021imap} and NICE-SLAM \citep{Zhu2022CVPR} being two real-time solutions.
The mapping runtime of such methods is weighed down by optimisation through the renderer.
Rendering can be sped up by decomposing parameters over space, e.g. by using voxels or primitives \citep{kilonerf,neuralvolumes,liu2020neural,mixtureprimitives,mueller2022instant}, but how to update neural maps in closed form remains unclear.
Therefore, we rely on vanilla voxel grid maps \citep{mirchev2021variational,occupancymaps}, as their probabilistic treatment and closed-form updates are straightforward, leaving implicit representations for future work.
We note that none of the aforementioned methods incorporate dynamics and uncertainty, which distinguishes our approach.
\paragraph{Probabilistic SLAM inference}
SLAM filters are thoroughly explored for flat 2D modelling \citep{probrobotics,kalman1960new,murphymap,fastslam,hahnel2003efficient,fastslam2,rbpfslam}, but have been superseded by MAP smoothing in modern visual SLAM (e.g. \citep{engel2014lsd,orbslam2,dso,vimo,vinsmono,Rosinol19icra-incremental,Rosinol20icra-Kimera}), primarily due to scalability concerns \citep{cadena2016past,strasdat2012visual}.
However, as of now smoothing is not computationally feasible without sparsity assumptions.
We therefore reexamine filtering for differentiable rendering, as we aim to obtain a dense map posterior with uncertainty in real-time (see \cref{app:motivation} for further motivation).
Filters may benefit from the dense modelling of observations \citep{strasdat2012visual}, which aligns with our objective, and we will demonstrate they can be a feasible solution for moderately-sized indoor environments.
For the states, we use a Laplace approximation \citep{laplace} and velocity updates similar to those in extended Kalman filters \citep{kalman1960new}.
For the map, occupancy grids are a common probabilistic choice \citep{occupancymaps,murphymap,bhm} and closed-form mapping has been used in that context \citep{rbpfslam}.
To enable rendering we provide a similar derivation, but for a signed distance function (SDF), which is related.
Probabilistic SDF mapping dates back to~\citet{curless1996volumetric}, and SDF updates have a well-known probabilistic interpretation \citep{hernandez2007probabilistic,dong2018psdf}.
We use these approximations to arrive at a holistic probabilistic solution that scales to dense 3D modelling in real-time.
\newcommand{\hist}{H}
\section{Overview} \label{sec:overview}
We approach on-line SLAM inference with two aims in mind.
First, we want to harmonise our map and state estimation with a predictive model.
Second, we want to quantify uncertainty: estimates and predictions should account for modelling inaccuracies as well as measurement and process noise.
Both are important for autonomous decision making.
To achieve this, we derive a Bayesian posterior in the probabilistic model of \citet{mirchev2021variational}, to ensure that inference matches the forward model.
Before we delve into our proposed solution, we present a practical summary.
At every time step:
\begin{enumerate}[topsep=0pt]
    \itemsep0em
    \item we point-estimate the agent's pose using gradient descent, involving geometry and dynamics.
    \item we extend the pose with a Gaussian covariance matrix through a Laplace approximation.
    \item with the pose, we estimate the agent's current velocity in closed form.
    \item with the pose and the current observation, we update the map in closed form.
\end{enumerate}
We use well-established methods for the above.
In 1. we combine assumed density filtering \citep{opper1999bayesian}, point-to-plane ICP \citep{chen1992object} and photometric alignment \citep{steinbrucker2011real,audras2011real}.
In 2. we use a Laplace approximation \citep{laplace,bishop}.
In 3. we use linear-Gaussian updates, akin to Kalman filters \citep{kalman1960new}.
In 4. we first derive generic closed-form map updates, which boil down to SDF updates \citep{curless1996volumetric} for our generative assumptions.

We contribute by deriving a holistic Bayesian inference from the generative model we started with.
In doing so, we identify where traditional techniques are applicable to make a practical algorithm.
\section{Methods}
In the following we will denote generative distributions, true posterior distributions and conditionals with $p(\cdot)$.
Respectively, approximate distributions will be denoted with $q(\cdot)$.
Approximation steps will be indicated by~$\approx$~in equations.
We use $q^{\varpars}\left(\cdot\right)$ to subsume estimated distribution parameters into $\varpars$.
A subscript $\cdot_t$ indicates that a variable or a distribution is different at every time step.

\subsection{Background}
We start with an overview of the generative model of \citet{mirchev2021variational} from which we will derive the inference.
We assume a sequence of RGB-D observations $\obs_{1:T}$ and a sequence of agent states $\state_{1:T}$ driven by controls $\control_{1:T-1}$ form a Markovian state-space model.
Each observation is constructed from a respective state with a rendering emission model $\pp{\obs_t}{\Map, \state_t}$, where $\Map$ is a global latent random variable for a dense map.
A transition model $\pp{\state_t}{\state_{t-1}, \control_{t-1}}$ accounts for the agent dynamics, where $\control_{t}$ are known acceleration controls.
Assuming $\state_1$ is given, the joint distribution is:
\vspace{-0.5em}
\eq{
    &\pp{\Map, \state_{2:T}, \obs_{1:T}}{\control_{1:T-1}, \state_1} = \p{\Map}\pp{\obs_1}{\Map, \state_1}\prod_{t=2}^{T} \pp{\state_t}{\state_{t-1}, \control_{t-1}} \pp{\obs_t}{\state_t, \Map}.
}
The map is a 3D voxel grid of occupancy and color--each cell contains four values.
The emission is fully-differentiable and performs volumetric raymarching, searching for a unique hit position at a surface along each ray \citep{parker1998interactive}.
The transition performs Euler integration, using the acceleration controls and maintained velocity from the latent state.
\Cref{app:generative} and the original paper have the details.

\subsection{Posterior Choice} \label{sec:posteriorchoice}
First we need to choose which posterior to approximate.
For example, \citet{mirchev2021variational} approximate the full posterior over the map and \emph{all} states $\pp{\Map, \state_{2:T}}{\obs_{1:T}, \control_{1:T-1}, \state_1}$ with variational inference~\citep{vae}.
While generic, this approach is slowed down by rendering at every optimisation step \citep{kayalibay2022tracking}, and the inevitable stochastic optimisation demands multiple steps until convergence.
In addition, estimating the posterior over all states scatters the optimisation budget across the whole trajectory.

To enable real-time inference we target an alternative posterior, the filter $\pp{\Map, \state_t}{\obs_{1:t}, \control_{1:t-1}, \state_1}$, as the last state belief is enough for planning ahead \citep{bertsekas}.
Since filters can be updated recursively \citep{sarkka2013bayesian,bishop}, we can use closed-form updates for fast inference.
Still, maintaining the joint distribution is too costly because of the large dense 3D map $\Map$.\footnote{E.g. the size of full-covariance Gaussian representations \citep{kalman1960new} or carrying multiple maps in parallel for a Rao-Blackwellised particle filter \citep{fastslam,grisetti2007improved} become prohibitive.}
Instead, we approximate the two marginal filters:
\eq{
   \qfilter{t}{\Map} &\approx \pfilter{t}{\Map} = \pp{\Map}{\obs_{1:t}, \control_{1:t-1}, \state_1} \\
    \qfilter{t}{\state_t} &\approx \pfilter{t}{\state_t} = \pp{\state_t}{\obs_{1:t}, \control_{1:t-1}, \state_1},
}
where $\hist_t = \historyplus$.
More details about this modelling choice can be found in \cref{app:marginals}.
We draw attention to the shorthand notation $\pfilter{t}{~\cdot}$, which will appear again in the following.

\subsection{Approximate Filtering}
For both marginal filters, we will arrive at adequate approximations by reusing the following equation:
\eq{
    \pp{\map, \state_t}{\hist_t} \propto&~ \pp{\obs_t}{\state_t, \map} \int \pp{\state_t}{\state_{t-1}, \control_{t-1}} \pp{\map, \state_{t-1}}{\hist_{t-1}} d\state_{t-1}, \numberthis \label{eq:chapmankolmogorov}
}
This is a classic recursive expression of the Bayes filter \citep{sarkka2013bayesian}.
Starting from each true marginal posterior, we will first expand the joint, then use \cref{eq:chapmankolmogorov} and apply a set of approximations.
Next we will discuss our final result, we defer the detailed derivation of both filters to \cref{app:mapfilter,app:statefilter}.

\subsubsection{Marginal Map Filter} \label{sec:mapfilter}
\def\mapupdate{\qq{\map}{\obs_t, \hat \state_t}}
We begin with the map approximation, starting from the true marginal Bayes filter:
\eq{
    \pp{\map}{\hist_t} =&~ \int \pp{\map, \state_t}{\hist_t} \dint\state_t \\
    \propto&~ \int \pp{\obs_t}{\state_t, \map} \int \pp{\state_t}{\state_{t-1}, \control_{t-1}} \pp{\map, \state_{t-1}}{\hist_{t-1}} \dint\state_{t-1} \dint\state_t \\
    \approx&~ \pp{\obs_t}{\hat \state_t, \map} \times \qfilter{t-1}{\map} \numberthis \label{eq:mapadf} \\
    \approx&~ \mapupdate \times \qfilter{t-1}{\map} =: \qfilter{t}{\map}. \numberthis \label{eq:mapupdate}
}
\Cref{eq:mapadf,eq:mapupdate} hide a few approximations detailed in \cref{app:mapfilter}.
The resulting solution takes a nominal state sample $\hat \state_t$, with which a map update $\mapupdate$ is applied to the previous map belief $\qfilter{t-1}{\Map}$.
We set $\hat \state_t$ to the mean of the current state belief $\qfilter{t}{\state_t}$.
Accepting some bias, we do this for speed as it is our best guess for $\state_t$ without extra computation.\footnote{\Cref{app:approximations} discusses this approximation further.}
Intuitively, the map update $\mapupdate$ populates the map such that the observation $\obs_t$ can be reconstructed.
Our derivation of the updates is similar to the one by \citet{rbpfslam} for 2D occupancy maps, but now applied to 3D. 

The above approximation is generic, agnostic to the specific map and rendering assumptions.
In practice, we need a closed-form map update $\mapupdate$ that is faithful to the emission $\pp{\obs_t}{\hat \state_t, \Map}$.
In this work, we follow \citet{mirchev2021variational} and use a Gaussian map that factorises over voxels:
\eq{
    \qfilter{t}{\map} = \prod_{ijk} \gauss{\map_{ijk}}{\bmu^\map_{ijk,t}, \diag((\bsigma^\map_{ijk,t})^2)}.
}
Here the indices $ijk$ run over voxels in a 3D grid.
For this specific representation and the assumed surface-based rendering, we identify that the map update $\mapupdate$ can be implemented as a probabilistic signed distance function (SDF) update \citep{curless1996volumetric}.
We provide the technical details in \cref{app:sdfs}.
SDF updates for voxel maps are a traditional concept in computer vision, and prior work has considered their probabilistic interpretation before \citep{hernandez2007probabilistic,dong2018psdf}.
We contribute by identifying the place of such updates in a probabilistic filter that follows the generative model of \citep{mirchev2021variational}.
A detailed discussion of how the above relates to classical SDF update equations can be found in \cref{app:sdfs}.

The above approximations are motivated by the real-time constraint.
For example, one could optimise \cref{eq:mapadf} directly with gradient descent through the renderer, but evaluating the emission is expensive and hinders accurate convergence on a budget.
This is particularly true when uncertainty estimates are desirable, as optimisation would then be stochastic and gradients noisy \citep{blei2017variational}.
In contrast, the derived one-shot map updates are meant to have a cost similar to emitting just once, while capturing uncertainty as well.
We show some of the differences between the two approaches in \cref{sec:runtime}.

\subsubsection{Marginal State Filter} \label{sec:statefilter}
\def\stateprior{\qqu{\state_t}{\control_{t-1}, \hist_{t-1}}{t}}
\def\poseprior{\qqu{\pose_t}{\control_{t-1}, \hist_{t-1}}{t}}
\def\velcond{\qqu{\vel_t}{\pose_t, \control_{t-1}, \hist_{t-1}}{t}}
Similarly, for the state filter we start from the true marginal and arrive at approximations via \cref{eq:chapmankolmogorov}:
\eq{
\pp{\state_t}{\hist_t}
=&~ \int \pp{\map, \state_t}{\hist_t} \dint\map \\
\propto&~ \int \pp{\obs_t}{\state_t, \map} \int \pp{\state_t}{\state_{t-1}, \control_{t-1}} \pp{\map, \state_{t-1}}{\hist_{t-1}} \dint\state_{t-1} \dint\map \\
\approx&~ \pp{\obs_t}{\pose_t, \hat \map} \poseprior \velcond \numberthis \label{eq:apx_objective} \\
\approx&~ \qfilter{t}{\pose_t} \times \velcond =:~ \qfilter{t}{\state_t} \numberthis \label{eq:stateupdate}.
}
We detail all the approximations that lead to \cref{eq:apx_objective} in \cref{app:statefilter}.
In \cref{eq:apx_objective} we have three terms: an image reconstruction likelihood, a Gaussian pose prior and a linear Gaussian velocity conditional given a pose.
The latter two we obtain analytically with a linear approximation of the transition model and the previous Gaussian belief $\qfilter{t-1}{\state_{t-1}}$ (\cf \cref{app:statefilter}).
First, using the first two terms of \cref{eq:apx_objective} we define a maximum a-posteriori (MAP) objective for pose optimisation:
\eq{
    \arg\max_{\pose_t}\,\, \log \pp{\obs_t}{\hat \Map, \pose_t} + \log \poseprior.
}
Here, $\hat \map$ is a nominal map sample set to the mean of the previous map belief $\qfilter{t-1}{\Map}$.\footnote{\Cref{app:approximations} discusses this approximation further.}
The term $\log \poseprior$ is an approximate dynamics prior over the current pose, it makes the pose respect the transition model.
The term $\log p(\obs_t\mid\hat \map, \pose_t)$ represents reconstructing the current observation, optimising it for the current pose will align the observation to the map.
However, evaluating this rendering term in every gradient step is inefficient.
Because of this, we replace it with the prediction-to-observation objective used by \citet{kayalibay2022tracking,niessner2013real,kinectfusion}.
We refer to \citep{kayalibay2022tracking} for further motivation and we list the technical details in \cref{app:statefilter}.

The above optimisation gives us a MAP pose estimate, which we denote with $\bmu^{\posetext}_t$.
Next, we apply a Laplace approximation \citep{laplace} around it to obtain a full covariance matrix $\boldsymbol{\Sigma}^{\posetext}_t$ which captures the curvature of the objective.
This leaves us with a full Gaussian belief over the current pose:
\eq{
    \qfilter{t}{\pose_t} = \gauss{\pose_t}{\bmu^{\posetext}_t, \boldsymbol{\Sigma}^{\posetext}_t}.
}
Finally, we can combine this Gaussian with the Gaussian velocity conditional $\velcond$ (the third term in \cref{eq:apx_objective}) into a full-state belief in closed form:
\eq{
    \qfilter{t}{\state_t} = \gauss{\state_t}{\bmu_t, \boldsymbol{\Sigma}_t} = \gauss{\pose_t}{\bmu^{\posetext}_t, \boldsymbol{\Sigma}^{\posetext}_t} \gauss{\vel_t}{\mathbf{D}_t\pose_t + \mathbf{e}_t, \boldsymbol{\Sigma^{\mathrm{vel}}_t}}.
}
This is approximate, we do it for speed and find it does not harm localisation in practice.
\Cref{app:statefilter} describes how the linear Gaussian terms come to be in more detail.
\section{Experiments} \label{sec:experiments}
\begin{figure}[t]
    \centering
    \begin{subfigure}[b]{0.45\textwidth}
        \centering
        \includegraphics[width=\linewidth]{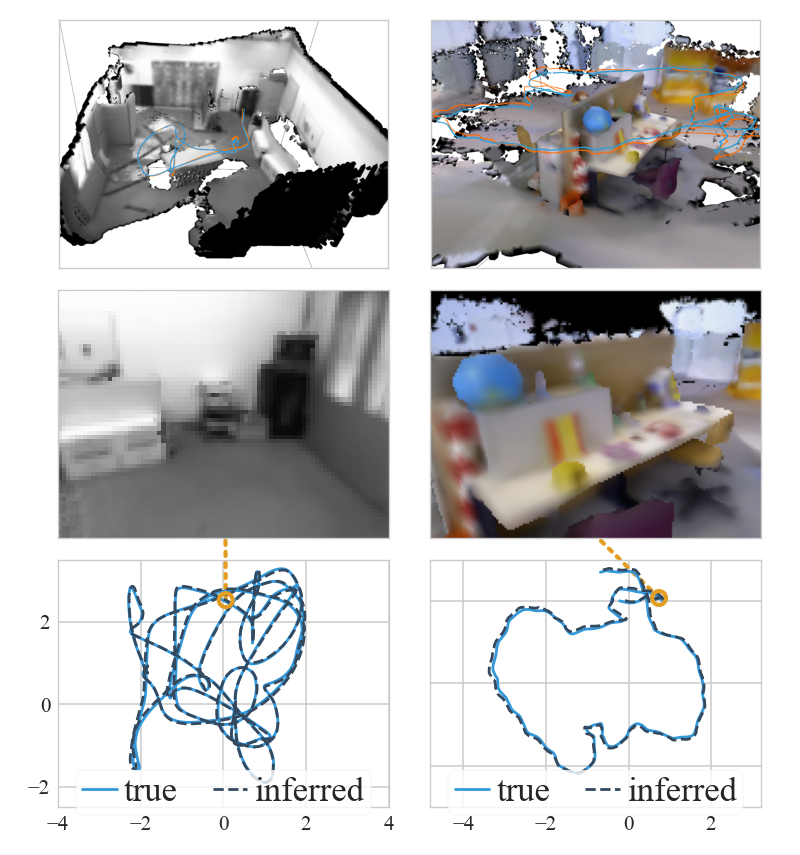}
        \caption{Example mapping and localisation}
        \label{fig:envs}
    \end{subfigure}%
    \begin{subfigure}[b]{0.45\textwidth}
            \centering
            \includegraphics[width=\linewidth]{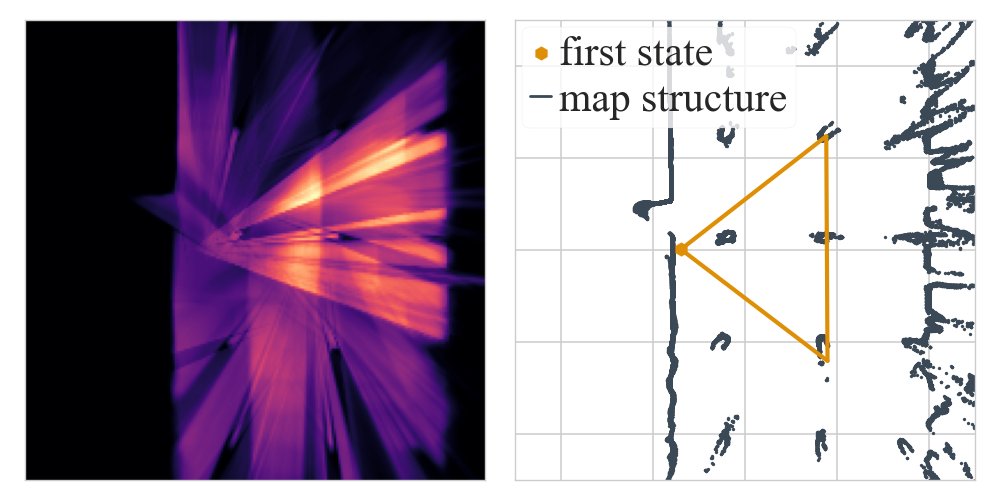}
            \caption{Blackbird map uncertainty.}
            \label{fig:mapuncertaintyblackbird}
            \includegraphics[width=\linewidth]{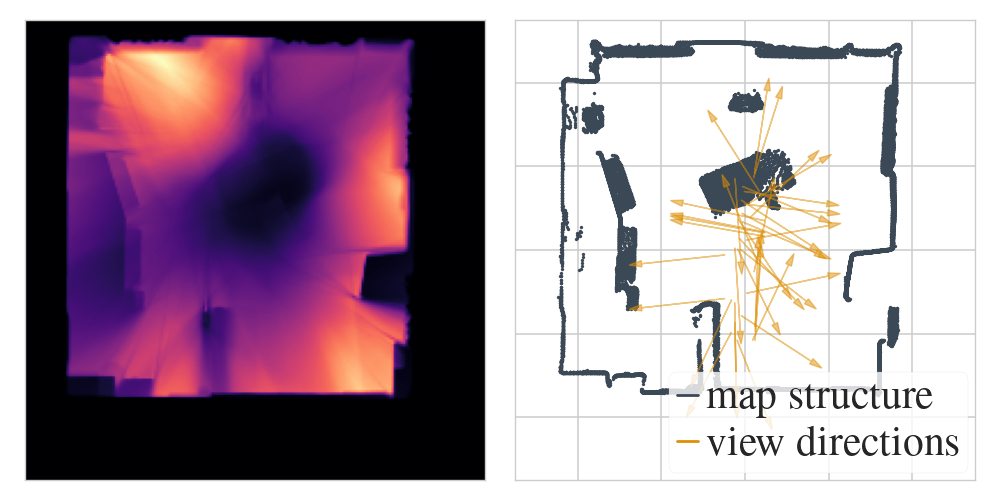}
            \caption{EuRoC map uncertainty.}
            \label{fig:mapuncertaintyeuroc}
    \end{subfigure}%
    \caption{
        (\subref{fig:envs}) 3D reconstruction, example emission and inferred trajectory for EuRoC/V102 and TUM-RGBD fr3/office.
        (\subref{fig:mapuncertaintyblackbird}) Blackbird experiment. Top-down map uncertainty on the left, \emph{black} is uncertain, \color{orange}\emph{orange}\color{black}\, is precise.
        Precision is highest in a triangle around the center, which is the camera frustum where the agent remains sitting on a platform for a long time, see the orange triangle amidst the map point cloud on the right.
        (\subref{fig:mapuncertaintyeuroc}) Analogous EuRoC experiment. Map uncertainty is high outside of the room, at the center and behind the two structures on the left due to occlusion.
        The uncertainty in the center is high because the agent primarily looks outwards (view directions in the right image).
    }
    \label{fig:bigpanel}
\end{figure}
Originally we set out with a few goals: the inference method should be faithful to the generative assumptions, it should quantify uncertainty and it should run in real-time.
What follows is an empirical analysis of these aspects.
We evaluate on the EuRoC \citep{euroc}, Blackbird  \citep{blackbird} and TUM-RGBD  \citep{tumrgbd} data sets.
The agent in the former two is an unmanned aerial vehicle (UAV), with speed of up to 4 m/s. 
For Blackbird, we use Semi-Global Block Matching (SGBM) for stereo depth estimation \citep{sgbm}.
For EuRoC, we use the ground-truth Leica MS50 depth readings provided by \citep{koestler2021tandem}.
We pretend the IMU readings from these data sets are our control inputs.
For TUM-RGBD we do not feed in any controls and assume a constant-velocity transition.
All experimental details are in \cref{app:experiments}.
\subsection{Inference Through a Probabilistic Generative Model}
First we look into the synergy between the inference and the generative assumptions.
In \cref{fig:envs} we see mapping and localisation examples.
The inferred scenes are consistent, with no dramatic offsets in geometry.
More importantly, rendering from the inferred map using the emission $\pp{\obs_t}{\state_t, \map}$ works as expected (see middle row), indicating that map updates are consistent with the generative assumptions.
This is evident from the accuracy of the inferred state trajectories as well (last row), as the pose optimisation objective from \cref{sec:statefilter} uses rendered images at every filtering step.
A potential discrepancy between the inference and the generative assumptions would lead to errors that would accumulate over time, which is not the case.
\begin{figure}
    \centering
    \begin{subfigure}{0.49\textwidth}
        \centering
        \includegraphics[width=\linewidth]{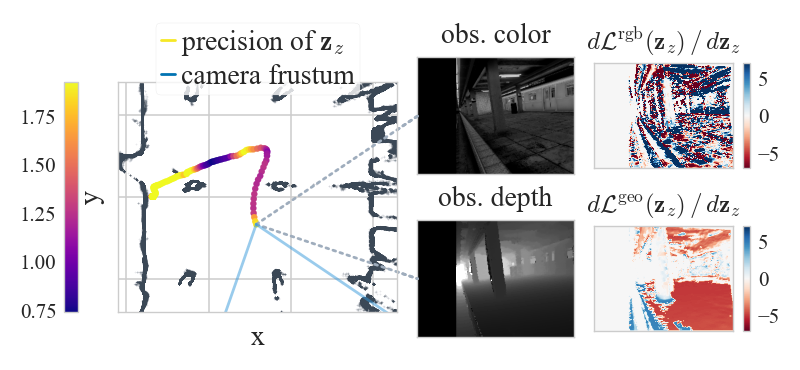}
        \caption{$z$ location uncertainty}
        \label{fig:stateuncertaintyx}
    \end{subfigure}
    \hfill
    \begin{subfigure}{0.49\textwidth}
        \centering
        \includegraphics[width=\linewidth]{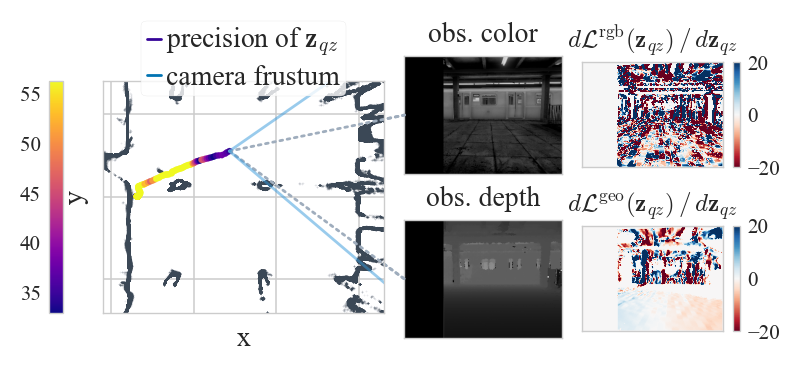}
        \caption{$\mathbf{q}_z$ (yaw) orientation uncertainty}
        \label{fig:stateuncertaintyqz}
    \end{subfigure}
    \caption{
        Inferred state uncertainty.
        Inferred trajectories are colored by precision (inverse uncertainty) of a certain state dimension, followed by observations, followed by columns of the tracking Jacobian for that same state dimension.
        (\subref{fig:stateuncertaintyx}) Here the precision in $z$ (vertical movement) is high (\color{Goldenrod}\emph{yellow}\color{black}),
        because the $z$-orthogonal floor produces a consistent Jacobian (bottom right).
        (\subref{fig:stateuncertaintyqz}) Here the precision in $\mathbf{q}_z$ orientation (yaw, azimuth) is low (\color{Violet}\emph{violet}\color{black}), as
        there are no orthogonal surfaces (\ie facing sideways).
        Note the low Jacobian magnitude of the horizontal floor this time (bottom right).
    }
    \label{fig:stateuncertainty}
\end{figure}
\paragraph{Map uncertainty} The inferred map uncertainty is determined by the map updates. 
We show its interpretable effects in \cref{fig:mapuncertaintyblackbird,fig:mapuncertaintyeuroc} for two examples, one from Blackbird and another from EuRoC.
Our map updates are akin to traditional SDF updates and the main factor that decides whether a map region is certain is how often it was observed.
Regions that were occluded by objects, are behind walls or were rarely in view remain uncertain, \eg as seen in \cref{fig:mapuncertaintyeuroc}.
In contrast, if the agent spends a lot of time looking at a certain map region, the uncertainty there decreases, as seen in \cref{fig:mapuncertaintyblackbird}.

\paragraph{State uncertainty} In \cref{fig:stateuncertainty} we analyse state uncertainty by looking at the variance for individual dimensions.
We notice that state uncertainty changes along the trajectory.
Uncertainty is determined by what the agent currently sees, based on the geometric relationship between the agent movement and the observed scene (\eg \cref{fig:stateuncertaintyx} and \cref{fig:stateuncertaintyqz}).
This effect can be explained if we examine the Laplace approximation used to estimate pose covariances.
At any given time step, we set the covariance to 
\eq{
    \boldsymbol{\Sigma}^{\posetext}_t \approx \mathbf{-H}^{-1} \approx -\left(2\mathbf{J}^{T}\mathbf{J}\right)^{-1}.
}
Here $\mathbf{H}$ is the Hessian of the tracking objective at the mean pose estimate and $\mathbf{J}$ is the Jacobian.
The Jacobian connects the pose to all image pixel errors.
The more consistent Jacobian entries are for a given pose dimension, the smaller the variance for that dimension will be.
We refer to \cref{app:uncertainty} for more details about the map and state uncertainty quality.

\subsection{Localisation Accuracy} \label{sec:localisation}
We compare PRISM's localisation to state-of-the-art methods in moderately-sized indoor environments.
We consider both baselines with dense maps (TANDEM \citep{koestler2021tandem}, VSSM-LM \citep{mirchev2021variational}, iMAP \citep{sucar2021imap}, NICE-SLAM \citep{Zhu2022CVPR}, CodeVIO \citep{codevio}) and sparse methods without rendering (ORB-SLAM2 \citep{orbslam2}, VINS \citep{vinsmono}, VIMO \citep{vimo}).
The results are in \cref{table:localisation}.
For the considered trajectories accuracy is comparable to the baselines, with differences of a few centimeters.
At the same time, our inference boasts a predictive state-space model with both rendering and dynamics as well as uncertainty estimates, which is not common in the dense visual SLAM literature.
Finally, in \cref{fig:velocities} we see example inferred agent velocities, noting the uncertainty bands.
This is possible because we model the agent dynamics.

Our localisation accuracy on Blackbird is better than the off-line variational inference results of VSSM-LM presented by \citet{mirchev2021variational}, and at the same time our solution runs in real-time and also captures uncertainty.
This shows the advantages of the proposed divide-and-conquer filtering.
\begin{table}[t]
    \centering
    \begin{minipage}{0.60\textwidth}
        \footnotesize
        \setlength{\tabcolsep}{2pt}
        \centering
        \captionof{table}{Localisation absolute error RMSE in meters on EuRoC \citep{euroc}, Blackbird \citep{blackbird} and TUM-RGBD \citep{tumrgbd}.}
        \label{table:localisation}
        \begin{tabular}{lcccc}
            \multirow{2}{*}{Trajectory}      &   \multirow{2}{*}{Ours} &    Code &   \multirow{2}{*}{TANDEM} &   ORB \\
                  &    &  VIO   &    &   SLAM2 \\
            \toprule
             EuRoC/V101 &              0.041 ($\pm$ 0.002)  &                 0.05 &       0.09 &             \textbf{0.031} \\
             EuRoC/V102 &              0.035 ($\pm$ 0.002) &                 0.07 &       0.17 &             \textbf{0.02} \\
             EuRoC/V103 &     \textbf{0.042 ($\pm$ 0.002)}  &                 0.07 &          - &             0.048 \\
             EuRoC/V201 &     \textbf{0.037 ($\pm$ 0.001)}  &                 0.10 &       0.09 &             \textbf{0.037} \\
             EuRoC/V202 &     \textbf{0.035 ($\pm$ 0.003)}  &                 0.06 &       0.12 &             \textbf{0.035} \\
             EuRoC/V203 &                                x &       \textbf{0.275} &          - &             x \\
            \bottomrule
            \addlinespace[1ex]
            \multirow{2}{*}{Trajectory}      &   \multirow{2}{*}{Ours} &    VSSM &   \multirow{2}{*}{VIMO} &   \multirow{2}{*}{VINS} \\
              & & LM & & \\
            \toprule
             picasso, 1 m/s &                   0.064 ($\pm$ 0.003) &     0.139 & \textbf{0.055} & 0.097 \\
             picasso, 2 m/s &                   0.053 ($\pm$ 0.003) &     0.136 & \textbf{0.040} & 0.043 \\
             picasso, 3 m/s &                   0.061 ($\pm$ 0.003) &     0.120 & \textbf{0.043} & 0.045 \\
             picasso, 4 m/s &               0.079 ($\pm$ 0.005)\tablefootnote{\label{blackbirdnote}Last  10 s are skipped, as the drone hits the ground during landing.} &     0.174 & \textbf{0.049} & 0.056 \\
             star, 1 m/s    &      0.089 ($\pm$ 0.007)\footref{blackbirdnote} &     0.137 &     \textbf{0.088} & 0.102 \\
             star, 2 m/s    &                   0.111 ($\pm$ 0.009) &     0.163 & \textbf{0.082} & 0.133 \\
             star, 3 m/s    &          \textbf{0.115 ($\pm$ 0.012)} &     0.281 &     0.183 & 0.235 \\
             star, 4 m/s    &          \textbf{0.153 ($\pm$ 0.015)}\footref{blackbirdnote} & 0.156 &         x &     x \\
            \bottomrule
            \addlinespace[1ex]
            \multirow{2}{*}{Trajectory}      &   \multirow{2}{*}{Ours} &    \multirow{2}{*}{iMAP} &   NICE &   ORB \\
               & & & SLAM & SLAM2$^*$ \\
            \toprule
            fr1/desk   &         0.053 ($\pm$ 0.003)  &     0.049 &            0.027 & \textbf{0.016} \\
            fr2/xyz    &         0.029 ($\pm$ 0.001)  &      0.02 &   \textbf{0.018} &           0.04 \\
            fr3/office &          0.083 ($\pm$ 0.001)  &     0.058 &             0.03 &  \textbf{0.01} \\
            \bottomrule
        \end{tabular}
    \end{minipage}
    \hfill
    \begin{minipage}{0.37\textwidth}
        \centering
        \includegraphics[width=1.0\linewidth]{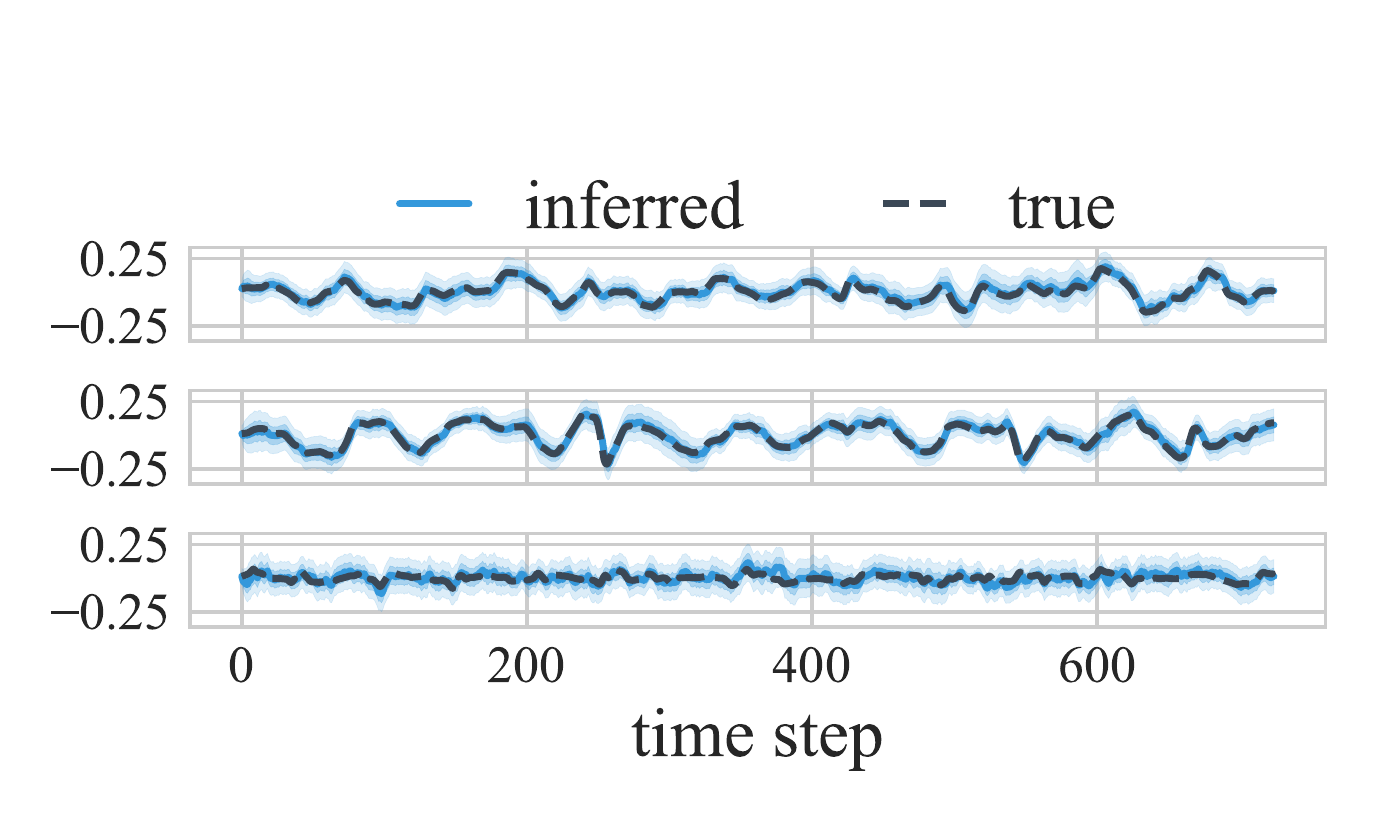}
        \captionof{figure}{Inferred $xyz$-velocity.}
        \vspace{-0.5em}
        \label{fig:velocities}
        \includegraphics[width=1.0\linewidth]{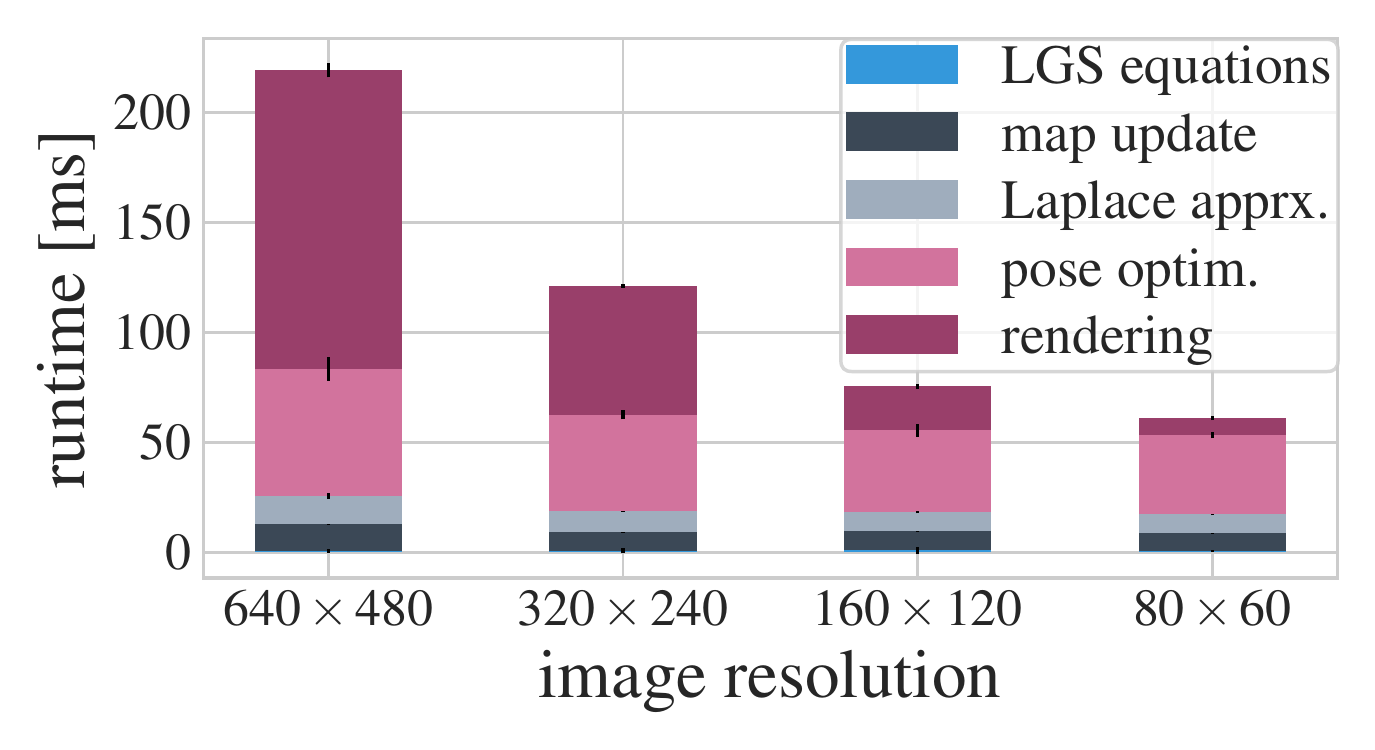}
        \captionof{figure}{Runtime breakdown.}
        \vspace{-0.5em}
        \label{fig:runtimes}
        \includegraphics[width=1.0\linewidth]{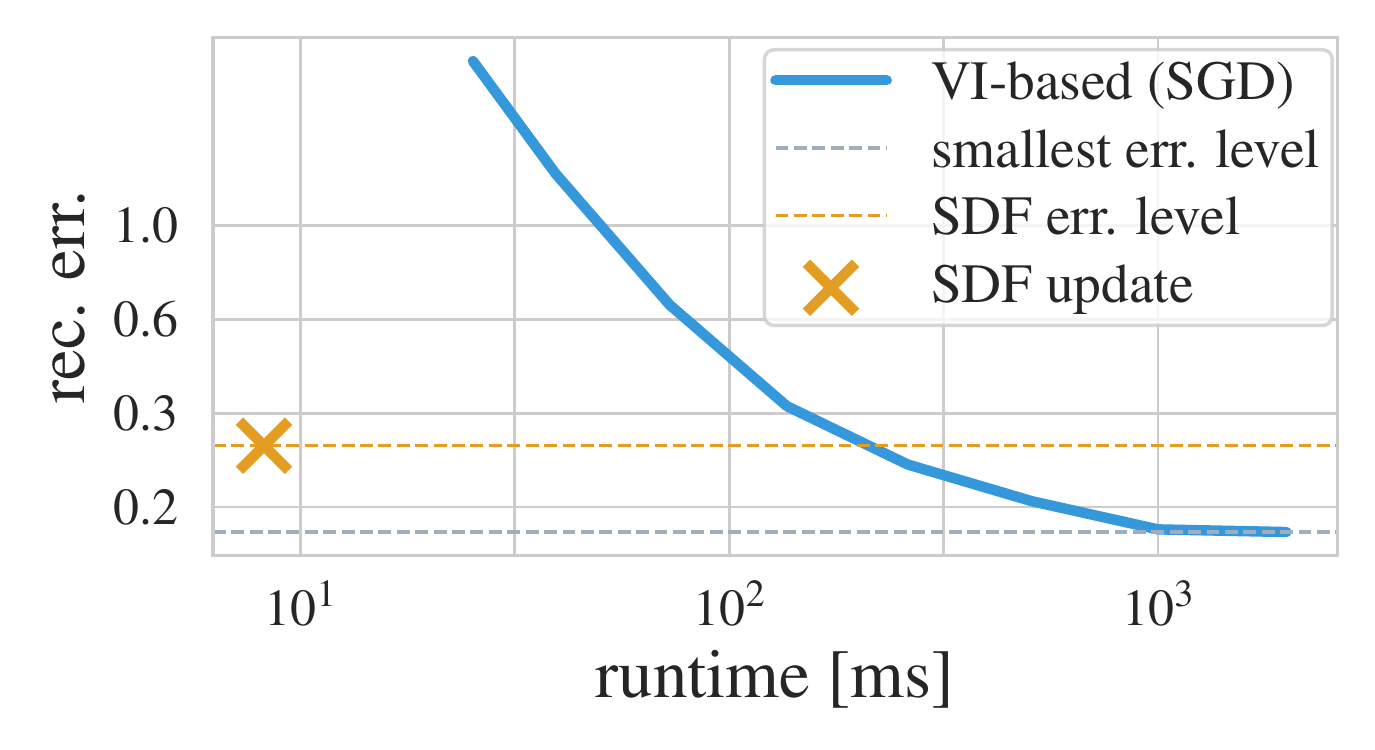}
        \captionof{figure}{Mapping comparison.}
        \vspace{-0.5em}
        \label{fig:sdfvsvi}
    \end{minipage}
\end{table}

\subsection{Approximations for Runtime Improvement} \label{sec:runtime}
All of our approximations are motivated by the real-time constraint, dictating the need for closed-form map updates, a Laplace approximation, linearisation assumptions and a surrogate pose optimisation objective.
\Cref{fig:runtimes} shows a runtime breakdown for different image resolutions, measured on an NVIDIA 1080 Ti GPU and an Intel(R) Xeon(R) W-2123 CPU at 3.6 GHz.
The heaviest operations are rendering and the gradient-based pose optimisation.
Based on movement speed, rendering can happen periodically, whenever a new anchor image prediction for pose optimisation is needed.
This leaves us with a total runtime of 10 Hz to 15 Hz, updating the map and state at every data step.
In \cref{fig:sdfvsvi} we also compare closed-form map updates to map inference via gradient-descent (e.g. as in \citep{mirchev2021variational,nerf,sucar2021imap,Zhu2022CVPR}).
While gradient-descent is more accurate on a bigger budget, it is much more expensive.
For example, to match the accuracy of the closed-form updates, which take less than 10 ms, one would need ca.\ 250 ms of optimisation, which is impractical.
These runtimes are for a voxel grid that is significantly faster than neural representations \citep{kayalibay2022tracking}, which would only exacerbate the problem.

\vspace{-0.5em}
\section{Limitations and Conclusion}
SDF voxel grids allow for closed-form updates, but their memory footprint limits the maximum resolution and scene size.
Voxel hashing~\citep{niessner2013real} or octrees \citep{steinbrucker2013large} can directly replace them for memory efficiency.
Neural maps and dynamically changing maps have remained out of our scope.
Their probabilistic formulation and closed-form updates require further investigation.
Our map factorises over voxels with no inter-region correlation, which could also be improved.
PRISM provides interpretable uncertainty in real-time, but estimation is approximate.
Obtaining perfectly calibrated uncertainty on a budget remains an open question (see \cref{app:uncertainty}).
While filtering works for our generative assumptions indoors, filters cannot revisit past errors and can drift in large scenes with high levels of exploration \citep{cadena2016past}.
We leave large-scale inference considerations for future work.

We have introduced PRISM, a method for probabilistic filtering in a predefined spatial state-space model.
Our solution runs in real-time, provides state and map uncertainty, and infers a dense map and a 6-DoF state trajectory with velocities.
It is comparably accurate to state-of-the-art SLAM in indoor environments.
To the best of our knowledge this is the first real-time fully-probabilistic solution for SLAM that combines differentiable rendering and agent dynamics.
We validated our method on three challenging data sets, featuring unmanned aerial vehicles and a handheld camera.
The results are promising, establishing PRISM as a viable state estimator for downstream model-based control.
 
\clearpage
\acknowledgments{We thank our reviewers for the thoughtful discussion, it helped us to better position our contribution.}

\subsubsection*{Notice}
This arXiv version of the paper is slightly adapted from the \href[]{https://proceedings.mlr.press/}{paper published at CoRL 2022, PMLR} by Atanas Mirchev, Baris Kayalibay, Ahmed Agha, Patrick van der Smagt, Daniel Cremers, Justin Bayer. The PMLR publication is licensed under \href[]{https://creativecommons.org/licenses/by/4.0/legalcode}{CC BY 4.0}.

\bibliography{related}

\begin{thebibliography}{91}
\providecommand{\natexlab}[1]{#1}
\providecommand{\url}[1]{\texttt{#1}}
\expandafter\ifx\csname urlstyle\endcsname\relax
  \providecommand{\doi}[1]{doi: #1}\else
  \providecommand{\doi}{doi: \begingroup \urlstyle{rm}\Url}\fi

\bibitem[Koller and Friedman(2009)]{koller2009probabilistic}
D.~Koller and N.~Friedman.
\newblock \emph{Probabilistic Graphical Models - Principles and Techniques}.
\newblock {MIT} Press, 2009.

\bibitem[Bertsekas(2005)]{bertsekas}
D.~P. Bertsekas.
\newblock \emph{Dynamic programming and optimal control, 3rd Edition}.
\newblock Athena Scientific, 2005.
\newblock ISBN 1886529264.
\newblock URL \url{http://www.worldcat.org/oclc/314894080}.

\bibitem[Engel et~al.(2014)Engel, Sch{\"o}ps, and Cremers]{engel2014lsd}
J.~Engel, T.~Sch{\"o}ps, and D.~Cremers.
\newblock Lsd-slam: Large-scale direct monocular slam.
\newblock In \emph{European conference on computer vision}, pages 834--849.
  Springer, 2014.

\bibitem[Mur-Artal and Tard{\'o}s(2017)]{orbslam2}
R.~Mur-Artal and J.~D. Tard{\'o}s.
\newblock Orb-slam2: An open-source slam system for monocular, stereo, and
  rgb-d cameras.
\newblock \emph{IEEE Transactions on Robotics}, 33\penalty0 (5):\penalty0
  1255--1262, 2017.

\bibitem[Engel et~al.(2018)Engel, Koltun, and Cremers]{dso}
J.~Engel, V.~Koltun, and D.~Cremers.
\newblock Direct sparse odometry.
\newblock \emph{IEEE Transactions on Pattern Analysis and Machine
  Intelligence}, Mar. 2018.

\bibitem[Cadena et~al.(2016)Cadena, Carlone, Carrillo, Latif, Scaramuzza,
  Neira, Reid, and Leonard]{cadena2016past}
C.~Cadena, L.~Carlone, H.~Carrillo, Y.~Latif, D.~Scaramuzza, J.~Neira, I.~Reid,
  and J.~J. Leonard.
\newblock Past, present, and future of simultaneous localization and mapping:
  Toward the robust-perception age.
\newblock \emph{IEEE Transactions on robotics}, 32\penalty0 (6):\penalty0
  1309--1332, 2016.

\bibitem[Kschischang et~al.(2001)Kschischang, Frey, and
  Loeliger]{kschischang2001factor}
F.~R. Kschischang, B.~J. Frey, and H.-A. Loeliger.
\newblock Factor graphs and the sum-product algorithm.
\newblock \emph{IEEE Transactions on information theory}, 47\penalty0
  (2):\penalty0 498--519, 2001.

\bibitem[Mildenhall et~al.(2020)Mildenhall, Srinivasan, Tancik, Barron,
  Ramamoorthi, and Ng]{nerf}
B.~Mildenhall, P.~P. Srinivasan, M.~Tancik, J.~T. Barron, R.~Ramamoorthi, and
  R.~Ng.
\newblock Nerf: Representing scenes as neural radiance fields for view
  synthesis.
\newblock In \emph{ECCV}, 2020.

\bibitem[Zhu et~al.(2022)Zhu, Peng, Larsson, Xu, Bao, Cui, Oswald, and
  Pollefeys]{Zhu2022CVPR}
Z.~Zhu, S.~Peng, V.~Larsson, W.~Xu, H.~Bao, Z.~Cui, M.~R. Oswald, and
  M.~Pollefeys.
\newblock Nice-slam: Neural implicit scalable encoding for slam.
\newblock In \emph{Proceedings of the IEEE/CVF Conference on Computer Vision
  and Pattern Recognition (CVPR)}, June 2022.

\bibitem[Koestler et~al.(2021)Koestler, Yang, Zeller, and
  Cremers]{koestler2021tandem}
L.~Koestler, N.~Yang, N.~Zeller, and D.~Cremers.
\newblock Tandem: Tracking and dense mapping in real-time using deep multi-view
  stereo.
\newblock In \emph{Conference on Robot Learning (CoRL)}, 2021.

\bibitem[Sucar et~al.(2021)Sucar, Liu, Ortiz, and Davison]{sucar2021imap}
E.~Sucar, S.~Liu, J.~Ortiz, and A.~J. Davison.
\newblock imap: Implicit mapping and positioning in real-time.
\newblock In \emph{2021 {IEEE/CVF} International Conference on Computer Vision,
  {ICCV} 2021, Montreal, QC, Canada, October 10-17, 2021}, pages 6209--6218.
  {IEEE}, 2021.

\bibitem[Kalman et~al.(1960)]{kalman1960new}
R.~E. Kalman et~al.
\newblock A new approach to linear filtering and prediction problems [j].
\newblock \emph{Journal of basic Engineering}, 82\penalty0 (1):\penalty0
  35--45, 1960.

\bibitem[Montemerlo et~al.(2002)Montemerlo, Thrun, Koller, and
  Wegbreit]{fastslam}
M.~Montemerlo, S.~Thrun, D.~Koller, and B.~Wegbreit.
\newblock Fastslam: A factored solution to the simultaneous localization and
  mapping problem.
\newblock In \emph{AAAI/IAAI}, 2002.

\bibitem[Grisetti et~al.(2007)Grisetti, Stachniss, and
  Burgard]{grisetti2007improved}
G.~Grisetti, C.~Stachniss, and W.~Burgard.
\newblock Improved techniques for grid mapping with rao-blackwellized particle
  filters.
\newblock \emph{IEEE transactions on Robotics}, 23\penalty0 (1):\penalty0
  34--46, 2007.

\bibitem[Mirchev et~al.(2021)Mirchev, Kayalibay, van~der Smagt, and
  Bayer]{mirchev2021variational}
A.~Mirchev, B.~Kayalibay, P.~van~der Smagt, and J.~Bayer.
\newblock Variational state-space models for localisation and dense 3d mapping
  in 6 dof.
\newblock In \emph{International Conference on Learning Representations}, 2021.
\newblock URL \url{https://openreview.net/forum?id=XAS3uKeFWj}.

\bibitem[S{\"a}rkk{\"a}(2013)]{sarkka2013bayesian}
S.~S{\"a}rkk{\"a}.
\newblock \emph{Bayesian Filtering and Smoothing}.
\newblock Number~3. Cambridge University Press, USA, 2013.
\newblock ISBN 1107619289.

\bibitem[Burri et~al.(2016)Burri, Nikolic, Gohl, Schneider, Rehder, Omari,
  Achtelik, and Siegwart]{euroc}
M.~Burri, J.~Nikolic, P.~Gohl, T.~Schneider, J.~Rehder, S.~Omari, M.~W.
  Achtelik, and R.~Siegwart.
\newblock The euroc micro aerial vehicle datasets.
\newblock \emph{The International Journal of Robotics Research}, 35\penalty0
  (10):\penalty0 1157--1163, 2016.

\bibitem[Antonini et~al.(2020)Antonini, Guerra, Murali, Sayre-McCord, and
  Karaman]{blackbird}
A.~Antonini, W.~Guerra, V.~Murali, T.~Sayre-McCord, and S.~Karaman.
\newblock The blackbird uav dataset.
\newblock \emph{The International Journal of Robotics Research}, 0\penalty0
  (0):\penalty0 0278364920908331, 2020.
\newblock \doi{10.1177/0278364920908331}.

\bibitem[Sturm et~al.(2012)Sturm, Engelhard, Endres, Burgard, and
  Cremers]{tumrgbd}
J.~Sturm, N.~Engelhard, F.~Endres, W.~Burgard, and D.~Cremers.
\newblock A benchmark for the evaluation of rgb-d slam systems.
\newblock In \emph{Proc. of the International Conference on Intelligent Robot
  Systems (IROS)}, Oct. 2012.

\bibitem[Rabiner(1989)]{hmms}
L.~Rabiner.
\newblock A tutorial on hidden markov models and selected applications in
  speech recognition.
\newblock \emph{Proceedings of the IEEE}, 77\penalty0 (2):\penalty0 257--286,
  1989.
\newblock \doi{10.1109/5.18626}.

\bibitem[Karl et~al.(2017)Karl, Soelch, Bayer, and van~der Smagt]{dvbf}
M.~Karl, M.~Soelch, J.~Bayer, and P.~van~der Smagt.
\newblock Deep variational bayes filters: Unsupervised learning of state space
  models from raw data.
\newblock In \emph{5th International Conference on Learning Representations,
  {ICLR} 2017, Toulon, France, April 24-26, 2017, Conference Track
  Proceedings}. OpenReview.net, 2017.

\bibitem[Krishnan et~al.(2015)Krishnan, Shalit, and Sontag]{dkf}
R.~G. Krishnan, U.~Shalit, and D.~Sontag.
\newblock Deep kalman filters.
\newblock \emph{arXiv preprint arXiv:1511.05121}, 2015.

\bibitem[Chung et~al.(2015)Chung, Kastner, Dinh, Goel, Courville, and
  Bengio]{vrnn}
J.~Chung, K.~Kastner, L.~Dinh, K.~Goel, A.~C. Courville, and Y.~Bengio.
\newblock A recurrent latent variable model for sequential data.
\newblock In C.~Cortes, N.~Lawrence, D.~Lee, M.~Sugiyama, and R.~Garnett,
  editors, \emph{Advances in Neural Information Processing Systems}, volume~28.
  Curran Associates, Inc., 2015.
\newblock URL
  \url{https://proceedings.neurips.cc/paper/2015/file/b618c3210e934362ac261db280128c22-Paper.pdf}.

\bibitem[Chiappa et~al.(2017)Chiappa, Racani{\`{e}}re, Wierstra, and
  Mohamed]{envsims}
S.~Chiappa, S.~Racani{\`{e}}re, D.~Wierstra, and S.~Mohamed.
\newblock Recurrent environment simulators.
\newblock In \emph{5th International Conference on Learning Representations,
  {ICLR} 2017, Toulon, France, April 24-26, 2017, Conference Track
  Proceedings}. OpenReview.net, 2017.
\newblock URL \url{https://openreview.net/forum?id=B1s6xvqlx}.

\bibitem[Ha and Schmidhuber(2018)]{worldmodels}
D.~Ha and J.~Schmidhuber.
\newblock Recurrent world models facilitate policy evolution.
\newblock In S.~Bengio, H.~Wallach, H.~Larochelle, K.~Grauman, N.~Cesa-Bianchi,
  and R.~Garnett, editors, \emph{Advances in Neural Information Processing
  Systems 31}, pages 2450--2462. Curran Associates, Inc., 2018.

\bibitem[Deisenroth and Rasmussen(2011)]{pilco}
M.~Deisenroth and C.~E. Rasmussen.
\newblock Pilco: A model-based and data-efficient approach to policy search.
\newblock In \emph{Proceedings of the 28th International Conference on machine
  learning (ICML-11)}, pages 465--472. Citeseer, 2011.

\bibitem[Hafner et~al.(2019)Hafner, Lillicrap, Fischer, Villegas, Ha, Lee, and
  Davidson]{planet}
D.~Hafner, T.~Lillicrap, I.~Fischer, R.~Villegas, D.~Ha, H.~Lee, and
  J.~Davidson.
\newblock Learning latent dynamics for planning from pixels.
\newblock In \emph{International Conference on Machine Learning}, pages
  2555--2565. PMLR, 2019.

\bibitem[Hafner et~al.(2020)Hafner, Lillicrap, Ba, and Norouzi]{dreamer}
D.~Hafner, T.~Lillicrap, J.~Ba, and M.~Norouzi.
\newblock Dream to control: Learning behaviors by latent imagination.
\newblock In \emph{International Conference on Learning Representations}, 2020.
\newblock URL \url{https://openreview.net/forum?id=S1lOTC4tDS}.

\bibitem[Lee et~al.(2020)Lee, Nagabandi, Abbeel, and Levine]{slac}
A.~X. Lee, A.~Nagabandi, P.~Abbeel, and S.~Levine.
\newblock Stochastic latent actor-critic: Deep reinforcement learning with a
  latent variable model.
\newblock \emph{Advances in Neural Information Processing Systems},
  33:\penalty0 741--752, 2020.

\bibitem[Becker{-}Ehmck et~al.(2020)Becker{-}Ehmck, Karl, Peters, and van~der
  Smagt]{learntofly}
P.~Becker{-}Ehmck, M.~Karl, J.~Peters, and P.~van~der Smagt.
\newblock Learning to fly via deep model-based reinforcement learning.
\newblock \emph{CoRR}, abs/2003.08876, 2020.

\bibitem[Karl et~al.(2019)Karl, Becker{-}Ehmck, Soelch, Benbouzid, van~der
  Smagt, and Bayer]{empowerment}
M.~Karl, P.~Becker{-}Ehmck, M.~Soelch, D.~Benbouzid, P.~van~der Smagt, and
  J.~Bayer.
\newblock Unsupervised real-time control through variational empowerment.
\newblock In \emph{Robotics Research - The 19th International Symposium {ISRR}
  2019, Hanoi, Vietnam, October 6-10, 2019}, volume~20 of \emph{Springer
  Proceedings in Advanced Robotics}, pages 158--173. Springer, 2019.

\bibitem[Corneil et~al.(2018)Corneil, Gerstner, and Brea]{vast}
D.~Corneil, W.~Gerstner, and J.~Brea.
\newblock Efficient model-based deep reinforcement learning with variational
  state tabulation.
\newblock volume~80 of \emph{Proceedings of Machine Learning Research}, pages
  1049--1058, Stockholmsmässan, Stockholm Sweden, 10--15 Jul 2018. PMLR.

\bibitem[Chua et~al.(2018)Chua, Calandra, McAllister, and Levine]{deepmbrl}
K.~Chua, R.~Calandra, R.~McAllister, and S.~Levine.
\newblock Deep reinforcement learning in a handful of trials using
  probabilistic dynamics models.
\newblock In \emph{Advances in Neural Information Processing Systems}, pages
  4754--4765, 2018.

\bibitem[Fraccaro et~al.(2018)Fraccaro, Rezende, Zwols, Pritzel, Eslami, and
  Viola]{gtmsm}
M.~Fraccaro, D.~J. Rezende, Y.~Zwols, A.~Pritzel, S.~M.~A. Eslami, and
  F.~Viola.
\newblock Generative temporal models with spatial memory for partially observed
  environments.
\newblock In \emph{Proceedings of the 35th International Conference on Machine
  Learning, {ICML} 2018, Stockholmsm{\"{a}}ssan, Stockholm, Sweden, July 10-15,
  2018}, volume~80 of \emph{Proceedings of Machine Learning Research}, pages
  1544--1553. {PMLR}, 2018.

\bibitem[Eslami et~al.(2018)Eslami, Jimenez~Rezende, Besse, Viola, Morcos,
  Garnelo, Ruderman, Rusu, Danihelka, Gregor, Reichert, Buesing, Weber,
  Vinyals, Rosenbaum, Rabinowitz, King, Hillier, Botvinick, Wierstra,
  Kavukcuoglu, and Hassabis]{gqn}
S.~M.~A. Eslami, D.~Jimenez~Rezende, F.~Besse, F.~Viola, A.~S. Morcos,
  M.~Garnelo, A.~Ruderman, A.~A. Rusu, I.~Danihelka, K.~Gregor, D.~P. Reichert,
  L.~Buesing, T.~Weber, O.~Vinyals, D.~Rosenbaum, N.~Rabinowitz, H.~King,
  C.~Hillier, M.~Botvinick, D.~Wierstra, K.~Kavukcuoglu, and D.~Hassabis.
\newblock Neural scene representation and rendering.
\newblock \emph{Science}, 360\penalty0 (6394):\penalty0 1204--1210, 2018.

\bibitem[Gregor et~al.(2019)Gregor, Rezende, Besse, Wu, Merzic, and van~den
  Oord]{gregor2019}
K.~Gregor, D.~J. Rezende, F.~Besse, Y.~Wu, H.~Merzic, and A.~van~den Oord.
\newblock Shaping belief states with generative environment models for rl.
\newblock In \emph{Advances in Neural Information Processing Systems}, pages
  13475--13487, 2019.

\bibitem[Mirchev et~al.(2019)Mirchev, Kayalibay, Soelch, van~der Smagt, and
  Bayer]{abise}
A.~Mirchev, B.~Kayalibay, M.~Soelch, P.~van~der Smagt, and J.~Bayer.
\newblock Approximate bayesian inference in spatial environments.
\newblock In \emph{Proceedings of Robotics: Science and Systems},
  FreiburgimBreisgau, Germany, June 2019.

\bibitem[Gupta et~al.(2020)Gupta, Tolani, Davidson, Levine, Sukthankar, and
  Malik]{gupta2019cognitive}
S.~Gupta, V.~Tolani, J.~Davidson, S.~Levine, R.~Sukthankar, and J.~Malik.
\newblock Cognitive mapping and planning for visual navigation.
\newblock \emph{Int. J. Comput. Vis.}, 128\penalty0 (5):\penalty0 1311--1330,
  2020.

\bibitem[Parisotto and Salakhutdinov(2018)]{parisotto2018neural}
E.~Parisotto and R.~Salakhutdinov.
\newblock Neural map: Structured memory for deep reinforcement learning.
\newblock In \emph{International Conference on Learning Representations}, 2018.
\newblock URL \url{https://openreview.net/forum?id=Bk9zbyZCZ}.

\bibitem[Chaplot et~al.(2020)Chaplot, Gandhi, Gupta, Gupta, and
  Salakhutdinov]{chaplot2020learning}
D.~S. Chaplot, D.~Gandhi, S.~Gupta, A.~Gupta, and R.~Salakhutdinov.
\newblock Learning to explore using active neural slam.
\newblock In \emph{International Conference on Learning Representations
  (ICLR)}, 2020.

\bibitem[Curless and Levoy(1996)]{curless1996volumetric}
B.~Curless and M.~Levoy.
\newblock A volumetric method for building complex models from range images.
\newblock In \emph{Proceedings of the 23rd annual conference on Computer
  graphics and interactive techniques}, pages 303--312, 1996.

\bibitem[Carr et~al.(2001)Carr, Beatson, Cherrie, Mitchell, Fright, McCallum,
  and Evans]{carr2001reconstruction}
J.~C. Carr, R.~K. Beatson, J.~B. Cherrie, T.~J. Mitchell, W.~R. Fright, B.~C.
  McCallum, and T.~R. Evans.
\newblock Reconstruction and representation of 3d objects with radial basis
  functions.
\newblock In \emph{Proceedings of the 28th annual conference on Computer
  graphics and interactive techniques}, pages 67--76, 2001.

\bibitem[Newcombe et~al.(2011{\natexlab{a}})Newcombe, Lovegrove, and
  Davison]{dtam}
R.~A. Newcombe, S.~Lovegrove, and A.~J. Davison.
\newblock Dtam: Dense tracking and mapping in real-time.
\newblock \emph{2011 International Conference on Computer Vision}, pages
  2320--2327, 2011{\natexlab{a}}.

\bibitem[Newcombe et~al.(2011{\natexlab{b}})Newcombe, Izadi, Hilliges,
  Molyneaux, Kim, Davison, Kohi, Shotton, Hodges, and
  Fitzgibbon]{newcombe2011kinectfusion}
R.~A. Newcombe, S.~Izadi, O.~Hilliges, D.~Molyneaux, D.~Kim, A.~J. Davison,
  P.~Kohi, J.~Shotton, S.~Hodges, and A.~Fitzgibbon.
\newblock Kinectfusion: Real-time dense surface mapping and tracking.
\newblock In \emph{2011 10th IEEE international symposium on mixed and
  augmented reality}, pages 127--136. IEEE, 2011{\natexlab{b}}.

\bibitem[Nie{\ss}ner et~al.(2013)Nie{\ss}ner, Zollh{\"o}fer, Izadi, and
  Stamminger]{niessner2013real}
M.~Nie{\ss}ner, M.~Zollh{\"o}fer, S.~Izadi, and M.~Stamminger.
\newblock Real-time 3d reconstruction at scale using voxel hashing.
\newblock \emph{ACM Transactions on Graphics (ToG)}, 32\penalty0 (6):\penalty0
  1--11, 2013.

\bibitem[Keller et~al.(2013)Keller, Lefloch, Lambers, Izadi, Weyrich, and
  Kolb]{keller2013realtime}
M.~Keller, D.~Lefloch, M.~Lambers, S.~Izadi, T.~Weyrich, and A.~Kolb.
\newblock Real-time 3d reconstruction in dynamic scenes using point-based
  fusion.
\newblock In \emph{2013 International Conference on 3D Vision - 3DV 2013},
  pages 1--8, 2013.
\newblock \doi{10.1109/3DV.2013.9}.

\bibitem[Whelan et~al.(2015)Whelan, Leutenegger, Salas-Moreno, Glocker, and
  Davison]{Whelan2015ElasticFusionDS}
T.~Whelan, S.~Leutenegger, R.~F. Salas-Moreno, B.~Glocker, and A.~J. Davison.
\newblock Elasticfusion: Dense slam without a pose graph.
\newblock In \emph{Robotics: Science and Systems}, 2015.

\bibitem[Cao et~al.(2018)Cao, Kobbelt, and Hu]{Cao2018RealtimeHT}
Y.-P. Cao, L.~P. Kobbelt, and S.~Hu.
\newblock Real-time high-accuracy three-dimensional reconstruction with
  consumer rgb-d cameras.
\newblock \emph{ACM Transactions on Graphics (TOG)}, 37:\penalty0 1 -- 16,
  2018.

\bibitem[Xu et~al.(2022)Xu, Nan, Zhou, Wang, and Wang]{hrbf}
Y.~Xu, L.~Nan, L.~Zhou, J.~Wang, and C.~C. Wang.
\newblock Hrbf-fusion: Accurate 3d reconstruction from rgb-d data using
  on-the-fly implicits.
\newblock \emph{ACM Transactions on Graphics (TOG)}, 41\penalty0 (3):\penalty0
  1--19, 2022.

\bibitem[Chen and Medioni(1992)]{chen1992object}
Y.~Chen and G.~Medioni.
\newblock Object modelling by registration of multiple range images.
\newblock \emph{Image and vision computing}, 10\penalty0 (3):\penalty0
  145--155, 1992.

\bibitem[Steinbr{\"u}cker et~al.(2011)Steinbr{\"u}cker, Sturm, and
  Cremers]{steinbrucker2011real}
F.~Steinbr{\"u}cker, J.~Sturm, and D.~Cremers.
\newblock Real-time visual odometry from dense rgb-d images.
\newblock In \emph{2011 IEEE international conference on computer vision
  workshops (ICCV Workshops)}, pages 719--722. IEEE, 2011.

\bibitem[Audras et~al.(2011)Audras, Comport, Meilland, and
  Rives]{audras2011real}
C.~Audras, A.~Comport, M.~Meilland, and P.~Rives.
\newblock Real-time dense appearance-based slam for rgb-d sensors.
\newblock In \emph{Australasian Conf. on Robotics and Automation}, volume~2,
  pages 2--2, 2011.

\bibitem[Kerl et~al.(2013)Kerl, Sturm, and Cremers]{kerl2013dense}
C.~Kerl, J.~Sturm, and D.~Cremers.
\newblock Dense visual slam for rgb-d cameras.
\newblock In \emph{2013 IEEE/RSJ International Conference on Intelligent Robots
  and Systems}, pages 2100--2106. IEEE, 2013.

\bibitem[Kayalibay et~al.(2022)Kayalibay, Mirchev, van~der Smagt, and
  Bayer]{kayalibay2022tracking}
B.~Kayalibay, A.~Mirchev, P.~van~der Smagt, and J.~Bayer.
\newblock Tracking and planning with spatial world models.
\newblock In \emph{Learning for Dynamics and Control Conference, {L4DC} 2022,
  23-24 June 2022, Stanford University, Stanford, CA, {USA}}, volume 168 of
  \emph{Proceedings of Machine Learning Research}, pages 124--137. {PMLR},
  2022.

\bibitem[Sitzmann et~al.(2020)Sitzmann, Martel, Bergman, Lindell, and
  Wetzstein]{siren}
V.~Sitzmann, J.~Martel, A.~Bergman, D.~Lindell, and G.~Wetzstein.
\newblock Implicit neural representations with periodic activation functions.
\newblock \emph{Advances in Neural Information Processing Systems}, 33, 2020.

\bibitem[Ortiz et~al.(2022)Ortiz, Clegg, Dong, Sucar, Novotn{\'{y}},
  Zollh{\"{o}}fer, and Mukadam]{isdf}
J.~Ortiz, A.~Clegg, J.~Dong, E.~Sucar, D.~Novotn{\'{y}}, M.~Zollh{\"{o}}fer,
  and M.~Mukadam.
\newblock isdf: Real-time neural signed distance fields for robot perception.
\newblock \emph{CoRR}, abs/2204.02296, 2022.

\bibitem[Lin et~al.(2021)Lin, Florence, Barron, Rodriguez, Isola, and
  Lin]{inerf}
Y.~Lin, P.~Florence, J.~T. Barron, A.~Rodriguez, P.~Isola, and T.~Lin.
\newblock inerf: Inverting neural radiance fields for pose estimation.
\newblock In \emph{{IEEE/RSJ} International Conference on Intelligent Robots
  and Systems, {IROS} 2021, Prague, Czech Republic, September 27 - Oct. 1,
  2021}, pages 1323--1330. {IEEE}, 2021.

\bibitem[Wang et~al.(2021)Wang, Wu, Xie, Chen, and Prisacariu]{nerfmm}
Z.~Wang, S.~Wu, W.~Xie, M.~Chen, and V.~A. Prisacariu.
\newblock Nerf-: Neural radiance fields without known camera parameters.
\newblock \emph{CoRR}, abs/2102.07064, 2021.

\bibitem[Reiser et~al.(2021)Reiser, Peng, Liao, and Geiger]{kilonerf}
C.~Reiser, S.~Peng, Y.~Liao, and A.~Geiger.
\newblock Kilonerf: Speeding up neural radiance fields with thousands of tiny
  mlps.
\newblock In \emph{2021 {IEEE/CVF} International Conference on Computer Vision,
  {ICCV} 2021, Montreal, QC, Canada, October 10-17, 2021}, pages 14315--14325.
  {IEEE}, 2021.

\bibitem[Lombardi et~al.(2019)Lombardi, Simon, Saragih, Schwartz, Lehrmann, and
  Sheikh]{neuralvolumes}
S.~Lombardi, T.~Simon, J.~Saragih, G.~Schwartz, A.~Lehrmann, and Y.~Sheikh.
\newblock Neural volumes.
\newblock \emph{ACM Transactions on Graphics}, 38\penalty0 (4):\penalty0
  1–14, Jul 2019.
\newblock ISSN 1557-7368.
\newblock \doi{10.1145/3306346.3323020}.
\newblock URL \url{http://dx.doi.org/10.1145/3306346.3323020}.

\bibitem[Liu et~al.(2020)Liu, Gu, Lin, Chua, and Theobalt]{liu2020neural}
L.~Liu, J.~Gu, K.~Z. Lin, T.-S. Chua, and C.~Theobalt.
\newblock Neural sparse voxel fields.
\newblock \emph{NeurIPS}, 2020.

\bibitem[Lombardi et~al.(2021)Lombardi, Simon, Schwartz, Zollhoefer, Sheikh,
  and Saragih]{mixtureprimitives}
S.~Lombardi, T.~Simon, G.~Schwartz, M.~Zollhoefer, Y.~Sheikh, and J.~Saragih.
\newblock Mixture of volumetric primitives for efficient neural rendering.
\newblock \emph{ACM Trans. Graph.}, 40\penalty0 (4), jul 2021.
\newblock ISSN 0730-0301.
\newblock \doi{10.1145/3450626.3459863}.
\newblock URL \url{https://doi.org/10.1145/3450626.3459863}.

\bibitem[M\"uller et~al.(2022)M\"uller, Evans, Schied, and
  Keller]{mueller2022instant}
T.~M\"uller, A.~Evans, C.~Schied, and A.~Keller.
\newblock Instant neural graphics primitives with a multiresolution hash
  encoding.
\newblock \emph{ACM Trans. Graph.}, 41\penalty0 (4):\penalty0 102:1--102:15,
  July 2022.
\newblock \doi{10.1145/3528223.3530127}.
\newblock URL \url{https://doi.org/10.1145/3528223.3530127}.

\bibitem[Moravec and Elfes(1985)]{occupancymaps}
H.~Moravec and A.~Elfes.
\newblock High resolution maps from wide angle sonar.
\newblock In \emph{Proceedings. 1985 IEEE International Conference on Robotics
  and Automation}, volume~2, pages 116--121. IEEE, 1985.

\bibitem[Thrun et~al.(2005)Thrun, Burgard, and Fox]{probrobotics}
S.~Thrun, W.~Burgard, and D.~Fox.
\newblock \emph{Probabilistic Robotics (Intelligent Robotics and Autonomous
  Agents)}.
\newblock The MIT Press, 2005.
\newblock ISBN 0262201623.

\bibitem[Murphy(1999)]{murphymap}
K.~P. Murphy.
\newblock Bayesian map learning in dynamic environments.
\newblock In \emph{Advances in Neural Information Processing Systems 12,
  {[NIPS} Conference, Denver, Colorado, USA, November 29 - December 4, 1999]},
  pages 1015--1021, 1999.

\bibitem[Hahnel et~al.(2003)Hahnel, Burgard, Fox, and
  Thrun]{hahnel2003efficient}
D.~Hahnel, W.~Burgard, D.~Fox, and S.~Thrun.
\newblock An efficient fastslam algorithm for generating maps of large-scale
  cyclic environments from raw laser range measurements.
\newblock In \emph{Proceedings 2003 IEEE/RSJ International Conference on
  Intelligent Robots and Systems (IROS 2003) (Cat. No.03CH37453)}, volume~1,
  pages 206--211 vol.1, 2003.
\newblock \doi{10.1109/IROS.2003.1250629}.

\bibitem[Montemerlo et~al.(2003)Montemerlo, Thrun, Koller, Wegbreit,
  et~al.]{fastslam2}
M.~Montemerlo, S.~Thrun, D.~Koller, B.~Wegbreit, et~al.
\newblock Fastslam 2.0: An improved particle filtering algorithm for
  simultaneous localization and mapping that provably converges.
\newblock In \emph{IJCAI}, volume~3, pages 1151--1156, 2003.

\bibitem[Grisetti et~al.(2005)Grisetti, Stachniss, and Burgard]{rbpfslam}
G.~Grisetti, C.~Stachniss, and W.~Burgard.
\newblock Improving grid-based slam with rao-blackwellized particle filters by
  adaptive proposals and selective resampling.
\newblock \emph{Proceedings of the 2005 IEEE International Conference on
  Robotics and Automation}, pages 2432--2437, 2005.

\bibitem[Nisar et~al.(2019)Nisar, Foehn, Falanga, and Scaramuzza]{vimo}
B.~Nisar, P.~Foehn, D.~Falanga, and D.~Scaramuzza.
\newblock Vimo: Simultaneous visual inertial model-based odometry and force
  estimation.
\newblock In \emph{Proceedings of Robotics: Science and Systems},
  FreiburgimBreisgau, Germany, June 2019.

\bibitem[Qin et~al.(2018)Qin, Li, and Shen]{vinsmono}
T.~Qin, P.~Li, and S.~Shen.
\newblock Vins-mono: A robust and versatile monocular visual-inertial state
  estimator.
\newblock \emph{IEEE Transactions on Robotics}, 34\penalty0 (4):\penalty0
  1004--1020, 2018.

\bibitem[Rosinol et~al.(2019)Rosinol, Sattler, Pollefeys, and
  Carlone]{Rosinol19icra-incremental}
A.~Rosinol, T.~Sattler, M.~Pollefeys, and L.~Carlone.
\newblock Incremental visual-inertial 3d mesh generation with structural
  regularities.
\newblock In \emph{2019 International Conference on Robotics and Automation
  (ICRA)}, 2019.

\bibitem[Rosinol et~al.(2020)Rosinol, Abate, Chang, and
  Carlone]{Rosinol20icra-Kimera}
A.~Rosinol, M.~Abate, Y.~Chang, and L.~Carlone.
\newblock Kimera: an open-source library for real-time metric-semantic
  localization and mapping.
\newblock In \emph{IEEE Intl. Conf. on Robotics and Automation (ICRA)}, 2020.
\newblock URL \url{https://github.com/MIT-SPARK/Kimera}.

\bibitem[Strasdat et~al.(2012)Strasdat, Montiel, and
  Davison]{strasdat2012visual}
H.~Strasdat, J.~M. Montiel, and A.~J. Davison.
\newblock Visual slam: why filter?
\newblock \emph{Image and Vision Computing}, 30\penalty0 (2):\penalty0 65--77,
  2012.

\bibitem[Laplace(1986)]{laplace}
P.~S. Laplace.
\newblock Memoir on the probability of the causes of events.
\newblock \emph{Statistical Science}, 1\penalty0 (3):\penalty0 364--378, 1986.
\newblock ISSN 08834237.
\newblock URL \url{http://www.jstor.org/stable/2245476}.

\bibitem[Senanayake and Ramos(2017)]{bhm}
R.~Senanayake and F.~Ramos.
\newblock Bayesian hilbert maps for dynamic continuous occupancy mapping.
\newblock In \emph{Proceedings of the 1st Annual Conference on Robot Learning},
  volume~78 of \emph{Proceedings of Machine Learning Research}, pages 458--471.
  PMLR, 13--15 Nov 2017.

\bibitem[Hern{\'a}ndez et~al.(2007)Hern{\'a}ndez, Vogiatzis, and
  Cipolla]{hernandez2007probabilistic}
C.~Hern{\'a}ndez, G.~Vogiatzis, and R.~Cipolla.
\newblock Probabilistic visibility for multi-view stereo.
\newblock In \emph{2007 IEEE Conference on Computer Vision and Pattern
  Recognition}, pages 1--8. IEEE, 2007.

\bibitem[Dong et~al.(2018)Dong, Wang, Wang, and Zha]{dong2018psdf}
W.~Dong, Q.~Wang, X.~Wang, and H.~Zha.
\newblock Psdf fusion: Probabilistic signed distance function for on-the-fly 3d
  data fusion and scene reconstruction.
\newblock In \emph{Proceedings of the European Conference on Computer Vision
  (ECCV)}, pages 701--717, 2018.

\bibitem[Opper and Winther(1999)]{opper1999bayesian}
M.~Opper and O.~Winther.
\newblock A bayesian approach to on-line learning.
\newblock 1999.

\bibitem[Bishop(2006)]{bishop}
C.~M. Bishop.
\newblock \emph{Pattern recognition and machine learning}.
\newblock springer, 2006.

\bibitem[Parker et~al.(1998)Parker, Shirley, Livnat, Hansen, and
  Sloan]{parker1998interactive}
S.~Parker, P.~Shirley, Y.~Livnat, C.~Hansen, and P.-P. Sloan.
\newblock Interactive ray tracing for isosurface rendering.
\newblock In \emph{Proceedings Visualization'98 (Cat. No. 98CB36276)}, pages
  233--238. IEEE, 1998.

\bibitem[Kingma and Welling(2014)]{vae}
D.~P. Kingma and M.~Welling.
\newblock Auto-encoding variational bayes.
\newblock In \emph{2nd International Conference on Learning Representations,
  {ICLR} 2014, Banff, AB, Canada, April 14-16, 2014, Conference Track
  Proceedings}, 2014.
\newblock URL \url{http://arxiv.org/abs/1312.6114}.

\bibitem[Blei et~al.(2017)Blei, Kucukelbir, and McAuliffe]{blei2017variational}
D.~M. Blei, A.~Kucukelbir, and J.~D. McAuliffe.
\newblock Variational inference: A review for statisticians.
\newblock \emph{Journal of the American statistical Association}, 112\penalty0
  (518):\penalty0 859--877, 2017.

\bibitem[Newcombe et~al.(2011)Newcombe, Izadi, Hilliges, Molyneaux, Kim,
  Davison, Kohli, Shotton, Hodges, and Fitzgibbon]{kinectfusion}
R.~A. Newcombe, S.~Izadi, O.~Hilliges, D.~Molyneaux, D.~Kim, A.~J. Davison,
  P.~Kohli, J.~Shotton, S.~Hodges, and A.~W. Fitzgibbon.
\newblock Kinectfusion: Real-time dense surface mapping and tracking.
\newblock \emph{2011 10th IEEE International Symposium on Mixed and Augmented
  Reality}, pages 127--136, 2011.

\bibitem[Hirschmuller(2007)]{sgbm}
H.~Hirschmuller.
\newblock Stereo processing by semiglobal matching and mutual information.
\newblock \emph{IEEE Transactions on pattern analysis and machine
  intelligence}, 30\penalty0 (2):\penalty0 328--341, 2007.

\bibitem[Zuo et~al.(2021)Zuo, Merrill, Li, Liu, Pollefeys, and Huang]{codevio}
X.~Zuo, N.~Merrill, W.~Li, Y.~Liu, M.~Pollefeys, and G.~Huang.
\newblock Codevio: Visual-inertial odometry with learned optimizable dense
  depth.
\newblock In \emph{2021 IEEE International Conference on Robotics and
  Automation (ICRA)}, pages 14382--14388. IEEE, 2021.

\bibitem[Steinbrucker et~al.(2013)Steinbrucker, Kerl, and
  Cremers]{steinbrucker2013large}
F.~Steinbrucker, C.~Kerl, and D.~Cremers.
\newblock Large-scale multi-resolution surface reconstruction from rgb-d
  sequences.
\newblock In \emph{Proceedings of the IEEE International Conference on Computer
  Vision}, pages 3264--3271, 2013.

\bibitem[Bradbury et~al.(2018)Bradbury, Frostig, Hawkins, Johnson, Leary,
  Maclaurin, Necula, Paszke, Vander{P}las, Wanderman-{M}ilne, and
  Zhang]{jax2018github}
J.~Bradbury, R.~Frostig, P.~Hawkins, M.~J. Johnson, C.~Leary, D.~Maclaurin,
  G.~Necula, A.~Paszke, J.~Vander{P}las, S.~Wanderman-{M}ilne, and Q.~Zhang.
\newblock {JAX}: composable transformations of {P}ython+{N}um{P}y programs,
  2018.
\newblock URL \url{http://github.com/google/jax}.

\bibitem[Kingma and Ba(2014)]{adam}
D.~P. Kingma and J.~Ba.
\newblock Adam: {A} method for stochastic optimization.
\newblock \emph{CoRR}, abs/1412.6980, 2014.

\bibitem[Jospin et~al.(2022)Jospin, Laga, Boussaid, Buntine, and
  Bennamoun]{jospin2022hands}
L.~V. Jospin, H.~Laga, F.~Boussaid, W.~Buntine, and M.~Bennamoun.
\newblock Hands-on bayesian neural networks—a tutorial for deep learning
  users.
\newblock \emph{IEEE Computational Intelligence Magazine}, 17\penalty0
  (2):\penalty0 29--48, 2022.

\bibitem[Sutton and Barto(2018)]{sutton2018reinforcement}
R.~S. Sutton and A.~G. Barto.
\newblock \emph{Reinforcement learning: An introduction}.
\newblock MIT press, 2018.

\end{thebibliography}

\newpage
\appendix
\section{The Cost of Maintaining a Joint Filter} \label{app:marginals}
In \cref{sec:posteriorchoice}, we mention that Rao-Blackwellised or full-covariance Gaussian representations for the joint $\pfilter{t}{\Map, \state_t}$ are difficult due to the large number of parameters in the dense 3D maps.
For example, a joint Gaussian distribution of $\Map$ and $\state_t$ would require $\mathcal{O}((n_\state + n_\Map)^2)$ parameters, where $n_\state$ is the size of a single state (6 degrees of freedom for the pose plus 6 degrees of freedom for the velocity) and $n_\Map$ is the size of a map (e.g. a voxel grid of size $200 \times 200 \times 200$).
Similarly, a Rao-Blackwellised particle representation would require $\mathcal{O}(P(n_\state + n_\Map))$, where $P \gg 1$ is a very large number of particles, leading to billions of parameters.
This is because every particle would carry its own individual map, and many particles are needed to properly cover the 6-DoF state space.
In these cases both the memory and the necessary computation to process all the data are prohibitive for real-time operation.

As a workaround, we choose to approximate the individual marginal distributions.
Framing the problem in this way helps with separation of concern and allows us to more easily incorporate traditional inference techniques (\cf \cref{sec:mapfilter,sec:statefilter} and \cref{app:mapfilter,app:statefilter}).
It is worth noting that the product of the two marginals $\qfilter{t}{\map}$ and $\qfilter{t}{\state_t}$ is not necessarily an optimal approximation of the joint.
As each approximate filter targets a marginal true posterior, their product is not directly optimised as a mean-field approximation \citep{bishop}.
The posterior approximation and particular derivation paths we have chosen is only one way (out of many) to frame the problem which we have found convenient to derive a practical solution.

One could attempt to sample from the joint, using one of the maintained marginal filters as a starting point.
For example, consider the following joint factorisation
\eq{
    \pfilter{t}{\Map, \state_t} &= \pfilter{t}{\Map}\pp{\state_t}{\Map, \hist_t}.
}
To sample from it (\eg\ via importance sampling \citep{bishop}), we would need to compute the term $\pp{\state_t}{\Map, \hist_t}$ up to a normalising constant:
\eq{
    \color{orange} \pp{\state_t}{\map, \hist_t} \color{black} &\propto \pp{\obs_t}{\Map, \state_t}\pp{\state_t}{\Map, \obs_{1:t-1}, \control_{1:t-1}} \\
    &= \pp{\obs_t}{\Map, \state_t} \int \pp{\state_t, \state_{t-1}}{\Map, \obs_{1:t-1}, \control_{1:t-1}} \dint \state_{t-1} \\
    &= \pp{\obs_t}{\Map, \state_t} \int \pp{\state_t}{\state_{t-1}, \control_{t-1}, \cancel{\Map, \obs_{1:t-1}, \control_{1:t-2}}} \\
    &\quad\qquad\qquad\qquad\qquad \times p(\state_{t-1} \mid \Map, \underbrace{\obs_{1:t-1}, \control_{1:t-2}}_{\hist_{t-1}}, \cancel{\control_{t-1}}) \dint \state_{t-1} \\
    &= \pp{\obs_t}{\Map, \state_t} \int \pp{\state_t}{\state_{t-1}, \control_{t-1}} \color{orange} \pp{\state_{t-1}}{\Map, \hist_{t-1}} \color{black} \dint \state_{t-1}. \numberthis \label{eq:joint_reweighting}\\
}
The above reveals a recursive expression for the evaluation of $\pp{\state_t}{\Map, \hist_t}$.
Therefore, computing this term would require going back the Markov chain to the beginning of the sequence, which is inefficient.
One could try to maintain cached approximations of $\pp{\state_t}{\Map, \hist_t}$ at every time step (\eg\ through weighted $\state_t$ particles, for different $\Map$), but this is encumbered by the large dimensionality of the map and states.
The alternative filter factorisation $\pfilter{t}{\Map, \state_t} = \pfilter{t}{\state_t}\pp{\Map}{\state_t, \hist_t}$ leads to an analogous problem.
Since we target real-time inference, we opt for maintaining approximations to only the marginal filters instead.
Further considerations about the joint posterior are left for future work.
\section{Details of the Generative Model} \label{app:generative}
We follow the generative assumptions of \citet{mirchev2021variational}.
The map prior is a 3D voxel grid of occupancy and color and factorises over voxels:
\eq{
    \p{\map} = \prod_{ijk} \gauss{\map_{ijk}}{\bmu_{ijk}, \sigma^2\mathbf{I})}.
}
Each map cell $\map_{ijk} \in \mathbb{R}^{4}$ contains an SDF value and three RGB values.
We assume an uninformed (very broad) prior, setting $\sigma \gg 1$.
We set the very first approximate map posterior to the prior (in the absence of data) and recusively apply updates to it as new data arrives.

Rendering from the map is captured by the emission model $\pp{\obs}{\Map, \state}$.
First, a bundle of rays are cast inside the camera frustum, using the pose $\pose$ to position them in space \footnote{$\obs$ is conditionally independent of $\vel$ given $\Map,\pose$.}.
Each ray point can be expressed as an offset from the camera center along a ray direction:
$$
\point^{ij} = \mathbf{c} + d \mathbf{r}^{ij}.
$$
Here $ij$ runs over pixels, and $d$ is the depth of the point and determines the length of the offset along the ray direction $\mathbf{r}^{ij}$.
For every ray, a discrete set of $K$ ray points $\{\point_k^{ij}\}_{[K]}$ is formed, using equidistantly spaced depth offsets $d \in \{\epsilon, 2\epsilon, 3\epsilon, \dots\}$.
For all ray points in the frustum, occupancy and color are obtained by evaluating the occupancy and color field $f_\Map: \mathbb{R}^3 \rightarrow \mathbb{R}^4$ parameterised by $\Map$.
This is done by trilinearly interpolating the cells of $\Map$.
Next, search is performed along each ray, finding the first point $\point_k^{ij}$ for which the occupancy exceeds a threshold $\tau$ (in our implementation $\tau = 0$).\footnote{We assume occupancy is continuous, deviating from the traditional $\{0, 1\}$ Bernoulli definition.}
This approximately finds the first intersection with a surface along the ray, but points along the ray are discrete.
To predict the depth of the surface more accurately, linear interpolation based on the occupancy values is used:
\eq{
    d^* = \alpha d_k + (1 - \alpha) d_{k-1},\qquad \alpha = \frac{\tau - f_{\mathrm{occ}}(\point^{ij}_{k - 1})}{f_{\mathrm{occ}}(\point^{ij}_{k}) - f_{\mathrm{occ}}(\point^{ij}_{k - 1})}.
}
Here $\point_{k -1}^{ij}$ is the point preceding the surface, and $\point_{k}^{ij}$ the point after it (identified by the ray search).

The linear interpolation of distance to the surface based on map content matches the assumptions of signed distance function representations (SDF) \citep{curless1996volumetric}.
If we consider a single ray in isolation, the SDF value would equal the signed distance of ray points to the surface, \ie predicted depth along the ray would be a linear function of the SDF values as well (identity), just like in the above equation.
This property of the generative renderer allows us to use closed-form map updates that are alike traditional SDF updates without sacrificing accuracy.
Therefore, we treat occupancy like an SDF with a flipped sign (positive inside objects, negative outside).
\Cref{app:sdfs} provides the details of the probabilistic map updates.

\def\rgbd{\mathrm{rgbd}}
Color predictions are evaluated analogously to depth for each ray.
The predicted depth and color are combined into an RGB-D image mean $\bmu_\rgbd$, which is used to parameterise a Laplace distribution:
\eq{
    \pp{\obs}{\Map, \state} = \mathrm{Laplace}(\obs \mid \bmu_\rgbd, \diag(\bsigma_E)).
}
The emission model is differentiable--$\Map$ and $\state$ can be optimised through it with gradient descent.
However, in this work we avoid using this gradient path, as it is too expensive for real-time inference.

The transition model $\pp{\state_{t+1}}{\state_t, \control_t}$ is defined as Euler integration of acceleration and velocity:
\eq{
    \pp{\state_{t+1}}{\state_t, \control_t} &= \gauss{\vel_{t+1}}{f_\veltext(\vel_t, \control_t), \diag(\bsigma^\veltext)^2} \\
    &\quad \times \gauss{\pose_{t+1}}{f_\posetext(\pose_t, \vel_{t+1}), \diag(\bsigma^\posetext)^2}.
}
Here $f_{\veltext}$ denotes acceleration integration and $f_{\posetext}$ denotes velocity integration.
We deviate from the formulation in the original paper, and use the velocity of step $t$ instead of $t-1$ to obtain the pose as step $t$.
We find this leads to a more convenient implementation when inferring velocity in a filtering setup.
We discuss this factorisation further in \cref{app:lintrans}, where we introduce the assumed linearisation of the transition.

We refer to the original paper for further details concerning the generative model.
\section{The Marginal Map Filter} \label{app:mapfilter}
Here we derive the approximation of the true marginal map filter:
\eq{
    \pfilter{t}{\Map} &= \int \pp{\map, \state_t}{\hist_t}\dint\state_t \\
        &\propto \int \pp{\obs_t}{\state_t, \map} \int \pp{\state_t}{\state_{t-1}, \control_{t-1}} \pp{\map, \state_{t-1}}{\hist_{t-1}} \dint\state_{t-1} \dint\state_t \\
        &\approx \int \pp{\obs_t}{\state_t, \map} \int \pp{\state_t}{\state_{t-1}, \control_{t-1}} \qfilter{t-1}{\Map}\qfilter{t-1}{\state_{t-1}} \dint\state_{t-1} \dint\state_t \numberthis \label{eq:apx_joint_marginals_map} \\
        &\approx \pp{\obs_t}{\hat \state_t, \map} \times \qfilter{t-1}{\map} \numberthis \label{eq:apx_nominal_state_map} \\
        &\approx \mapupdate \times \qfilter{t-1}{\map} =:~\qfilter{t}{\map} \numberthis \label{eq:apx_update_map}.
}
We begin by applying \cref{eq:chapmankolmogorov} directly.
The first approximation we make is in \cref{eq:apx_joint_marginals_map}, substituting the true previous joint filter $\pfilter{t-1}{\map, \state_{t-1}}$ for $\qfilter{t-1}{\Map}\qfilter{t-1}{\state_{t-1}}$.
We do this for speed, sacrificing some modelling accuracy for the sake of directly reusing the previous marginal estimates.
\Cref{app:marginals} discusses why using an approximation of the joint here is difficult.
Next, in \cref{eq:apx_nominal_state_map} we use a single-sample MC approximation of the integral of $\state_t$.
The nominal value $\hat \state_t$ we set to the mean of the current approximate state filter $\qfilter{t}{\state_t}$.
Accepting some bias, we also do this for speed, as this is the best guess for $\state_t$ available at step $t$ without extra computation.
Our next approximation is the term $\mapupdate$ in \cref{eq:apx_update_map}, which represents a closed-form map update.
Intuitively, it approximately inverts the emission $\pp{\obs_t}{\hat \state_t, \Map}$ and populates the map with the content necessary for reconstructing the observation $\obs_t$.

To understand how the map update comes to be, consider the following:
\eq{
    \pp{\obs_t}{\hat \state_t, \Map} &= \pp{\obs_t, \Map}{\hat \state_t}\p{\Map}^{-1} \\
    &= \pp{\Map}{\obs_t, \hat \state_t}\underbrace{\pp{\obs_t}{\hat \state_t}}_{\text{const in $\Map$}}\p{\Map}^{-1} \\
    &= \pp{\Map}{\obs_t, \hat \state_t}\times \p{\Map}^{-1} \times \mathrm{const} \\
    &\approx \qq{\Map}{\obs_t, \hat \state_t},
}
where $\qq{\Map}{\obs_t, \hat \state_t}$ has to be designed to be proportional to $\pp{\Map}{\obs_t, \hat \state_t}\times \p{\Map}^{-1}$.
With the help of Bayes' theorem we invert the emission and then arrive at an expression which is the target for the map update approximation.
The same strategy has been previously used by \citet{rbpfslam}.
We specify the exact form of the update in \cref{app:sdfs}, where we engineer the update such that it results in meaningful rendering.

\section{Map Update Formulation} \label{app:sdfs}
Consider the emission model $\pp{\obs_t}{\state_t, \map}$.
As described in \cref{app:generative}, it determines the hit of a single surface during raycasting.
Moreover, the linear interpolation described in \cref{app:generative} means an SDF-like representation will match the rendering assumptions.
Noting that we use an uninformed map prior, the map update then needs to approximate $\qq{\Map}{\obs_t, \hat \state_t} \approx \pp{\Map}{\obs_t, \hat \state_t} \approx \pp{\Map}{\obs_t, \hat \state_t}\times \p{\Map}^{-1}$.
Because of this, we define the map update as:
\def\bLambda{\boldsymbol{\Lambda}}
\def\mapupdateprec{\bLambda^{\mathrm{update}}}
\def\mapfilterprec{\bLambda^{\map}_t}
\def\mapfilterprecplus{\bLambda^\map_{t+1}}
\def\mapupdatemean{f_{\mathrm{update}}(\state_t, \obs_t)_{ijk}}
\def\mapfiltermean{\bmu^\map_{ijk,t}}
\def\mapupdatecov{\diag(\bsigma_{ijk}^{\mathrm{update}})^2}
\def\mapfiltercov{\diag(\bsigma^\map_{ijk,t})^2}
\def\mapupdategauss{\gauss{\Map_{ijk}}{\mapupdatemean, \mapupdatecov}}
\def\mapfiltergauss{\gauss{\map_{ijk}}{\mapfiltermean, \mapfiltercov}}
\eq{
    \qq{\Map}{\obs_t, \state_t} &= \prod_{ijk} \qq{\Map_{ijk}}{\obs_t, \state_t} \\
    \qq{\Map_{ijk}}{\obs_t, \state_t} &= \mapupdategauss \\
    f_{\mathrm{update}}(\state_t, \obs_t)_{ijk} &= [-f_\sdf(\point_{ijk}, \state_t, \obs_t), f_\rgb(\point_{ijk}, \state_t, \obs_t)]^T.
}
Here the indices $ijk$ run over the voxels in a 3D grid.
The function $f_\sdf$ computes the SDF update values for the particular voxel center $\point_{ijk}$ based on the observed depth image $\obs_t^d$ and the camera pose $\pose_t$.
$f_\rgb$ computes the RGB update values analogously.
Both functions follow a traditional implementation~\citep{curless1996volumetric}.
We use the negative SDF value in the update to match the assumptions of \citet{mirchev2021variational} (see \cref{app:generative}).
Since the update is engineered in advance to match the generative assumptions, we empirically validate whether it is appropriate in \cref{sec:experiments}.

Next we recap the parametric form of the approximate map filter:
\eq{
    \qfilter{t}{\map} = \prod_{ijk} \mapfiltergauss.
}
Applying the update is straightforward, as it comes down to solving the multiplication of Gaussians for each voxel in closed form.
This is because both the update and the approximate map filter factorise over voxels.
Therefore, the application of the updates can be easily parallelised on a GPU.
In the following we present the update equations for the whole map at once for the sake of simplicity of notation:
\eq{
    \qq{\map}{\obs_t, \state_t}\qfilter{t}{\map} &= \gauss{\map}{f_{\mathrm{update}}(\state_t, \obs_t), \diag(\bsigma^{\mathrm{update}})^2} \\
    &\quad \times \gauss{\map}{\bmu^\map_t, \diag(\bsigma_{t}^\map)^2} \\
    &= \gauss{\map}{\bmu^\map_{t+1}, \diag(\bsigma_{t + 1}^\map)^2} := \qfilter{t+1}{\map}. \numberthis \label{eq:mapupdategauss}
}
The update equations are better described in terms of the Gaussian precisions (inverse covariances), denoting them as:
\eq{
    \mapfilterprec = \diag(\bsigma^\map_t)^{-2} \\
    \mapfilterprecplus = \diag(\bsigma^\map_{t+1})^{-2} \\
    \mapupdateprec = \diag(\bsigma^{\mathrm{update}})^{-2}.
}
Solving for the parameters $\bmu_{t+1}^\map$ and $\mapfilterprecplus$, from \cref{eq:mapupdategauss} we have:
\eq{
    \bmu_{t+1}^\map &= (\mapfilterprecplus)^{-1} \left(\mapfilterprec \bmu^\map + \mapupdateprec f_{\mathrm{update}}(\state_t, \obs_t)\right) \\
    \mapfilterprecplus &= \mapfilterprec + \mapupdateprec.
}
These equations reveal the connection to the traditional SDF equations of~\citet{curless1996volumetric}:
\eq{
    D_{t+1}(\point) &= \frac{W_t(\point)D_t(\point) + w_t(\point) d_t(\point)}{W_t(\point) + w_t(\point)} \\
    W_{t+1}(\point) &= W_t(\point) + w_t(\point).
}
Here $\point \in \mathbb{R}^3$ is a point in the world frame (\eg\ a voxel center), $D_t$ is the accumulated SDF, $d_t$ is the SDF update, $W_t$ the accumulated weights so far and $w_t$ the update weight.
The algebraic form is the same, equating the mean of the filtering estimate to $D$, the mean of the update to $d$, the precision of the filtering estimate to $W$ and the precision of the update to $w$.
A similar probabilistic connection has been explored before in \citep{hernandez2007probabilistic,dong2018psdf}.

\section{The Marginal State Filter} \label{app:statefilter}
Before discussing the state filter, first we need to introduce the following linear approximation of the transition.

\subsection{Transition Linearisation} \label{app:lintrans}
We represent each state $\state_t = (\pose_t, \vel_t)$ as the combination of a pose $\pose \in \SE$ and a velocity (translational and angular) $\vel \in \mathbb{R}^6$.
Poses are parameterised as a combination of a 3D location and a quaternion. 
Our controls are translational and angular acceleration in the world reference frame: $\control_t = (\control^{\text{lin. accel}}_t, \control^{\text{ang. accel}}_t)$.
The assumed generative transition model is then the Euler integration of acceleration first, and then of velocity:
\eq{
    \pp{\state_{t+1}}{\state_t, \control_t} &= \pp{\vel_{t+1}}{f_\veltext(\vel_t, \control_t)}\pp{\pose_{t+1}}{f_\posetext(\pose_t, \vel_{t+1})} \\
    &= \gauss{\vel_{t+1}}{f_\veltext(\vel_t, \control_t), \diag(\bsigma^\veltext)^2} \\
    &\quad \times \gauss{\pose_{t+1}}{f_\posetext(\pose_t, \vel_{t+1}), \diag(\bsigma^\posetext)^2}.
}
The function $f_{\veltext}$ defines acceleration integration in the world frame and is linear:
\eq{
    \vel_{t+1} = f_\veltext(\vel_t, \control_t) = \vel_t + \control_t \cdot (\Delta t)^2.
}
The function $f_{\posetext}$ defines Euler integration of the agents velocity, to obtain its new pose.
When we do inference (not for prediction), we choose to linearise this function:
\eq{
    \pose_{t+1} &= f_\posetext(\pose_t, \vel_{t+1}) \\
        &\approx \underbrace{\mathbf{A}_t \pose_t + \mathbf{B}_t \vel_{t+1} + \mathbf{c}_t}_{\text{first order Taylor apprx.}} =:~ \hat f_{\posetext}(\pose_t, \vel_{t+1}).
}
This assumption lets us propagate uncertainty and infer velocity in closed form.
Thus, we define an approximate linearised transition:
\eq{
    \qq{\state_{t+1}}{\state_t, \control_t} &\approx \pp{\state_{t+1}}{\state_t, \control_t} \numberthis \label{eq:lintrans} \\
    \qq{\state_{t+1}}{\state_t, \control_t} &= \pp{\vel_{t+1}}{f_\veltext(\vel_t, \control_t)}\qq{\pose_{t+1}}{\hat f_\posetext(\pose_t, \vel_{t+1})} \\
    &= \gauss{\vel_{t+1}}{\vel_t + \control_t \cdot (\Delta t)^2, \diag((\bsigma^\veltext)^2)} \\
    &\quad \times \gauss{\pose_{t+1}}{\mathbf{A}_t \pose_t + \mathbf{B}_t \vel_{t+1} + \mathbf{c}_t, \diag((\bsigma^\posetext)^2)}.
}
Here, we introduce a linearisation $\hat f_\posetext$ of conventional Euler pose integration.
Since the velocity integration function $f_\veltext$ is linear by definition, this leaves us with two linear Gaussian conditionals, which we will use for closed-form updates \citep{bishop}.
In particular, the following integral can be solved in closed form:
\eq{
    \int \qq{\state_t}{\state_{t-1}, \control_{t-1}} \qfilter{t-1}{\state_{t-1}} \dint\state_{t-1} =: \qqu{\state_t}{\control_{t-1}, \hist_{t-1}}{t} \numberthis \label{eq:statepropagation},
}
assuming $\qfilter{t-1}{\state_{t-1}}$ is a Gaussian belief over the previous state, our state filter approximation introduced in \cref{app:statefilterderiv}.

\subsection{State Filter Derivation} \label{app:statefilterderiv}
We can now derive the approximation of the true marginal state filter:
\eq{
    \pfilter{t}{\state_t} &= \int \pp{\map, \state_t}{\hist_t}\dint\map \\
        &\propto \int \pp{\obs_t}{\state_t, \map} \int \pp{\state_t}{\state_{t-1}, \control_{t-1}} \pfilter{t-1}{\map, \state_{t-1}} \dint\state_{t-1} \dint\map\\
        &\approx \int \pp{\obs_t}{\state_t, \map} \int \qq{\state_t}{\state_{t-1}, \control_{t-1}} \qfilter{t-1}{\map}\qfilter{t-1}{\state_{t-1}} \dint\state_{t-1} \dint\map \numberthis \label{eq:apx_joint_marginals_state} \\
        &\approx \pp{\obs_t}{\state_t, \hat \map} \int \qq{\state_t}{\state_{t-1}, \control_{t-1}} \qfilter{t-1}{\state_{t-1}} \dint\state_{t-1} \numberthis \label{eq:apx_nominal_map_state} \\
        &= \pp{\obs_t}{\pose_t, \hat \map} \qqu{\state_t}{\control_{t-1}, \hist_{t-1}}{t} \numberthis \label{eq:apx_linear_state} \\
        &= \pp{\obs_t}{\pose_t, \hat \map} \qqu{\pose_t}{\control_{t-1}, \hist_{t-1}}{t} \qqu{\vel_t}{\pose_t, \control_{t-1}, \hist_{t-1}}{t} \numberthis \label{eq:apx_vel_post_state} \\
        &\approx \qfilter{t}{\pose_t} \qqu{\vel_t}{\pose_t, \control_{t-1}, \hist_{t-1}}{t} =:~ \qfilter{t}{\state_t} \numberthis \label{eq:apx_pose_opt_state}.
}
Our first approximation is in \cref{eq:apx_joint_marginals_state}, substituting $\pp{\map, \state_{t-1}}{\hist_{t-1}}$ for $\qfilter{t-1}{\map}\qfilter{t-1}{\state_{t-1}}$ with the same reasoning as for the map filter.
We also replace the true transition model for the linearised version from \cref{eq:lintrans}.
Next, in \cref{eq:apx_nominal_map_state} we MC-estimate the integral of $\map$ with a single sample, using the mean $\hat \map$ of the previous map belief $\qfilter{t}{\map}$.
Next, in \cref{eq:apx_linear_state} we solve the integral over $\state_{t-1}$ analytically (\cf\ \cref{eq:statepropagation}).
We can do this because the approximate linear transition $\qq{\state_t}{\state_{t-1}, \control_{t-1}}$ forms a linear Gaussian system with $\qfilter{t-1}{\state_{t-1}}$.
Thus we obtain $\qqu{\state_t}{\control_{t-1}, \hist_{t-1}}{t}$, an approximate Gaussian prior over the current state.
Next, in \cref{eq:apx_vel_post_state} we can split this Gaussian into $\poseprior$, a Gaussian prior over the current agent pose, and $\velcond$, a linear Gaussian velocity conditional given a pose.
We can obtain both of them in closed form following standard multivariate Gaussian equations for linear Gaussian systems \citep{bishop}.

\subsection{Pose Optimisation Objective}
As already discussed in the main text, we obtain a pose belief $\qfilter{t}{\pose_t}$ using a maximum-aposteriori (MAP) objective
\eq{
    \arg\max_{\pose_t}\,\, \log \pp{\obs_t}{\hat \Map, \pose_t} + \log \qqu{\pose_t}{\control_{t-1}, \hist_{t-1}}{t}.
}
It arises naturally from the first two terms in \cref{eq:apx_vel_post_state}, serving as a likelihood and a prior.
The term $\log\pp{\obs_t}{\hat \map, \pose_t}$ represents rendering with the emission model, and evaluating it in every gradient step is inefficient.
Because of this, we replace that term for the prediction-to-observation objective used by \citet{kayalibay2022tracking,niessner2013real,kinectfusion}.
We refer to \citep{kayalibay2022tracking} for a discussion of why this surrogate is meaningful.
Our final objective for optimising the pose is:
\eq{
    &\arg\min_{\pose_t}\,\, \loss[\mathrm{geo}]{\pose_t, \obs_t, \pose_{t-1}, \hat{\obs}_{t-1}} + \loss[\rgb]{\pose_t, \obs_t, \pose_{t-1}, \hat{\obs}_{t-1}} \\
    &\quad - \log \qqu{\pose_{t}}{\control_{t-1}, \hist_{t-1}}{t} \numberthis \label{eq:pointtoplanelaplace}.
}
where we have:
\eq{
    \loss[\mathrm{geo}]{\pose_t, \obs_t, \pose_{t-1}, \hat{\obs}_{t-1}} &= \sum_k \norm{\langle \hat\point_{t-1}^k- \transform_{\pose_{t}}^{\pose_{t-1}}\point_{t}^k, \hat\normal_{t-1}^k \rangle}_1 \\
    \loss[\rgb]{{\pose_t, \obs_t, \pose_{t-1}, \hat{\obs}_{t-1}}} &= \sum_k \norm{\hat\obs_{t-1}^\rgb[\pi(\transform_{\pose_{t}}^{\pose_{t-1}}\point_{t}^k)] - \obs_{t}^\rgb[\pi(\point_{t}^k)]}_1.
}
We follow the notation of \citet{kayalibay2022tracking} for consistency, and refer to that paper for further details.
$\pose_t$ is the current unknown pose of the agent.
$\pose_{t-1}$ is the mean pose of the previous pose belief.
$\hat{\obs}_{t-1}$ is a rendered observation from the preceding step, the mean of $\pp{\obs_{t-1}}{\hat \Map, \pose_{t-1}}$.
$\transform_{\pose_{t}}^{\pose_{t-1}}$ is the relative pose between the current and previous data step.
$\point_{t}^k$ and $\hat\point_{t-1}^k$ are corresponding 3D points in respectively the current and previous camera frames.
Accordingly, $\hat\normal_{t-1}^k$ is the corresponding normal of the rendered depth image.

The two terms $\loss[\mathrm{geo}]$ and $\loss[\mathrm{\rgb}]$ align the current RGB-D observation $\obs_t$ to the preceding prediction $\hat \obs_{t-1}$ rendered from the map, using geometric and photometric alignment, also known as point-to-plane ICP \citep{curless1996volumetric} and direct color image alignment \citep{audras2011real,steinbrucker2011real}.
Because the preceding $\hat \obs_{t-1}$ is rendered, this effectively anchors the current observation to the map by optimising the new pose of the agent.
In addition, $\log \qqu{\pose_t}{\control_{t-1}, \hist_{t-1}}{t}$ is maximized, satisfying the linearised dynamics prior over the agent pose.
The L1 norm in $\loss[\mathrm{geo}]$ and $\loss[\mathrm{\rgb}]$ corresponds to a Laplace distribution assumption and the sole reason for it is robustness to outliers.
When we apply a Laplace approximation\footnote{Not to be confused with a Laplace distribution assumption.} to obtain pose uncertainty, we implicitly change the L1 assumption.
This is because we approximate a Hessian for the objective with the square of the full-objective Jacobian (with the dynamics prior term as well), which implies an assumed curvature of a square function.
In practice we did not observe any major negative consequences from this fact.
\section{Approximation Gap} \label{app:approximations}
In our derivations, we have made a few crude approximations to reduce computations and make implementation simpler.
For example, in \cref{eq:apx_nominal_state_map}, we choose to use the previous state filter's mean instead of taking the expectation \wrt the whole distribution when computing the map marginal filter.
Similarly, in \cref{eq:apx_nominal_map_state}, we choose to use the previous map filter's mean instead of taking the expectation \wrt the whole distribution when computing the state marginal filter.
Note that the implication of this conditioning is that state uncertainty is not reflected in the map updates (\ie map does not become more uncertain if we are uncertain where to place the update) and map uncertainty is not reflected in the pose optimisation (\ie states do not become more uncertain even if the map for which they are optimised has not settled yet).
We look forward to improving this aspect in future work, as proper uncertainty propagation can stabilise long-term operation of the proposed inference and allow for better overall uncertainty calibration of the model.
In \cref{eq:apx_joint_marginals_map,eq:apx_joint_marginals_state} we also approximate the joint posterior with the product of both marginal approximations.
We believe that positioning the method as an approximation to $\pfilter{t}{\state_t}$ and $\pfilter{t}{\Map}$, the optimal marginal posteriors, reveals the exact places where compromises have been made.
We expect that explicitly highlighting the current approximation gap will be conducive for future research.

\section{Experiment Details} \label{app:experiments}
\subsection{Execution Details}
We have implemented PRISM with JAX \citep{jax2018github}, using Accelerated Linear Algebra (XLA) to compile computations into kernels that can be executed on a GPU device.
This lets us execute everything in real-time, while preserving the auto-differentiability of the generative model \citep{mirchev2021variational}.
Rendering, map updates and the Laplace approximation for pose uncertainty are executed on GPU, as they involve a lot of parallel computations.
The gradient-based pose optimisation is executed on CPU, as every optimisation step is lightweight and optimisation steps need to happen in sequence.
The linear-Gaussian updates are executed on CPU as well.

\subsection{Hyperparameters and Inference Details}
\paragraph{Map parameters}
Grid resolution is $200 \times 200 \times 200$ across all experiments.
The map is parameterised with a mean grid and a grid with standard deviations.
Each mean grid cell contains occupancy and RGB color, four values in total.
Each standard deviation grid cell also contains four respective uncertainty values.
Each map covers a hypercube of real-world space, we list the environment sizes per data set in \cref{table:envsizes}.
This determines the effective voxel size, between 7 cm for EuRoC and TUM-RGBD and 12.5 cm for Blackbird.
\paragraph{Transition parameters}
The transition is homoscedastic, with predefined scales for acceleration and velocity integration.
We use different scales for the location and orientation components of the states.
\Cref{table:transscales} provides the hyperparameters, with $\bsigma^\veltext$ governing acceleration integration and $\bsigma^\posetext$ governing velocity integration.
They are the same for all data sets.
\begin{table}
    \centering
    \caption{Assumed environment sizes, one size per data set.}
    \vspace{1em}
    \label{table:envsizes}
    \begin{tabular}[b]{ccc}
        \toprule
        \textbf{EuRoC} & \textbf{Blackbird} & \textbf{TUM-RGBD} \\
        \midrule
        14 m $\times$ 14 m $\times$ 14 m & 25 m $\times$ 25 m $\times$ 25 m & 14 m $\times$ 14 m $\times$ 14 m \\
        \bottomrule
    \end{tabular}
    \vspace{1em}
    \caption{Transition scale hyperparameters.}
    \label{table:transscales}
    \begin{tabular}[b]{cccc}
        \toprule
        \multicolumn{2}{c}{\textbf{translation}} & \multicolumn{2}{c}{\textbf{rotation}} \\
         $\bsigma^\veltext$ & $\bsigma^\posetext$ & $\bsigma^\veltext$ & $\bsigma^\posetext$ \\
        \midrule
        0.03 & 0.05 & 0.03 & 0.02 \\
        \bottomrule
    \end{tabular}
\end{table}
\paragraph{Rendering}
The maximum camera depth is set to 7.0 m for EuRoC, 8.0 m for TUM-RGBD and 20.0 m for Blackbird.
The ray step size $\epsilon$ is set to $0.4 \times \mathrm{voxel\_size}$.
The threshold for hit determination $\tau$ is set to 0, so that rendering is compatible with the map updates.
\paragraph{Map updates}
Map updates are placed only for map voxels that fall inside the camera frustum, and fall between the camera optical center and an added truncation distance after the observed depth surface.
The truncation distance is $4 \times \mathrm{voxel\_size}$ for Blackbird \citep{blackbird} and $2 \times \mathrm{voxel\_size}$ for EuRoC \citep{euroc} and TUM-RGBD \citep{tumrgbd}.
We assume a constant scale $\sigma_{ijk}^{\mathrm{update}} = 1.0$ for the update of all relevant voxels.
For the very first map belief, we initialise occupancy (negative SDF) to -0.001 and color to 0.0 for all map voxels.
\paragraph{Pose optimisation}
Poses are optimised using the MAP objective from \cref{sec:statefilter}.
We use Adam \citep{adam} as an optimiser, disabling its momentum.
Step sizes are set individually for the translation and rotation components of the optimised poses.
For the translational part of the pose, we use a step size of 0.001.
For the rotational part of the pose, we use a step size of 0.00036.
We use 1000 optimisation steps, sampling 200 random pixels uniformly at each step to evaluate the objective.
Pixels with a geometric error higher than 0.45 are ignored during optimisation.
Pixels with a photometric error higher than 0.15 are ignored during optimisation.
The same optimisation hyperparameters are used in all experiments.

In terms of the objective itself, we assume a different Laplace scale for the photometric and geometric terms in \cref{eq:pointtoplanelaplace} for each data set, based on our confidence in the sensors.
A lower Laplace scale means higher priority is given to the respective observations (color or depth).
For EuRoC and TUM-RGBD, we use a color scale of 0.1 and a geometric scale of 0.02, as the depth readings in these data sets are accurate.
For Blackbird, we use a color scale of 0.02 and a geometric scale of 0.2, as the depth we use in Blackbird is rather inaccurate, estimated with SGBM \citep{sgbm}.
\paragraph{Laplace approximation}
We approximate the pose optimisation objective's Hessian with the square product of the objective's Jacobian.
Since the Laplace estimates are noisy over time due to approximation errors, we apply an exponential moving average over time with a coefficient of 0.8.

\subsection{Data Preprocessing}
We subsample images to a resolution of $60 \times 80$ pixels for EuRoC, $192 \times 256$ for Blackbird and $120 \times 160$ pixels for TUM-RGBD.
Since the ground-truth depth readings for EuRoC from \citep{koestler2021tandem} are sparse we downscale them to a resolution of 60 $\times$ 80 pixels to densify.
We ignore pixels with invalid depth throughout our method, as well as pixels for which depth is highly discontinuous.

\subsection{Localisation Evaluation Details}
We compare our localisation results to the published results of existing SLAM methods, carrying them over from the respective publications.
Respectively, the choice of our evaluation trajectories was determined by whether a comparison is possible.
We run inference experiments with 5 random seeds for each trajectory and report the mean and standard deviation of the relevant metrics.
\newpage
\section{Uncertainty Analysis} \label{app:uncertainty}
\begin{figure}[H]
    \centering
    \includegraphics[width=\linewidth]{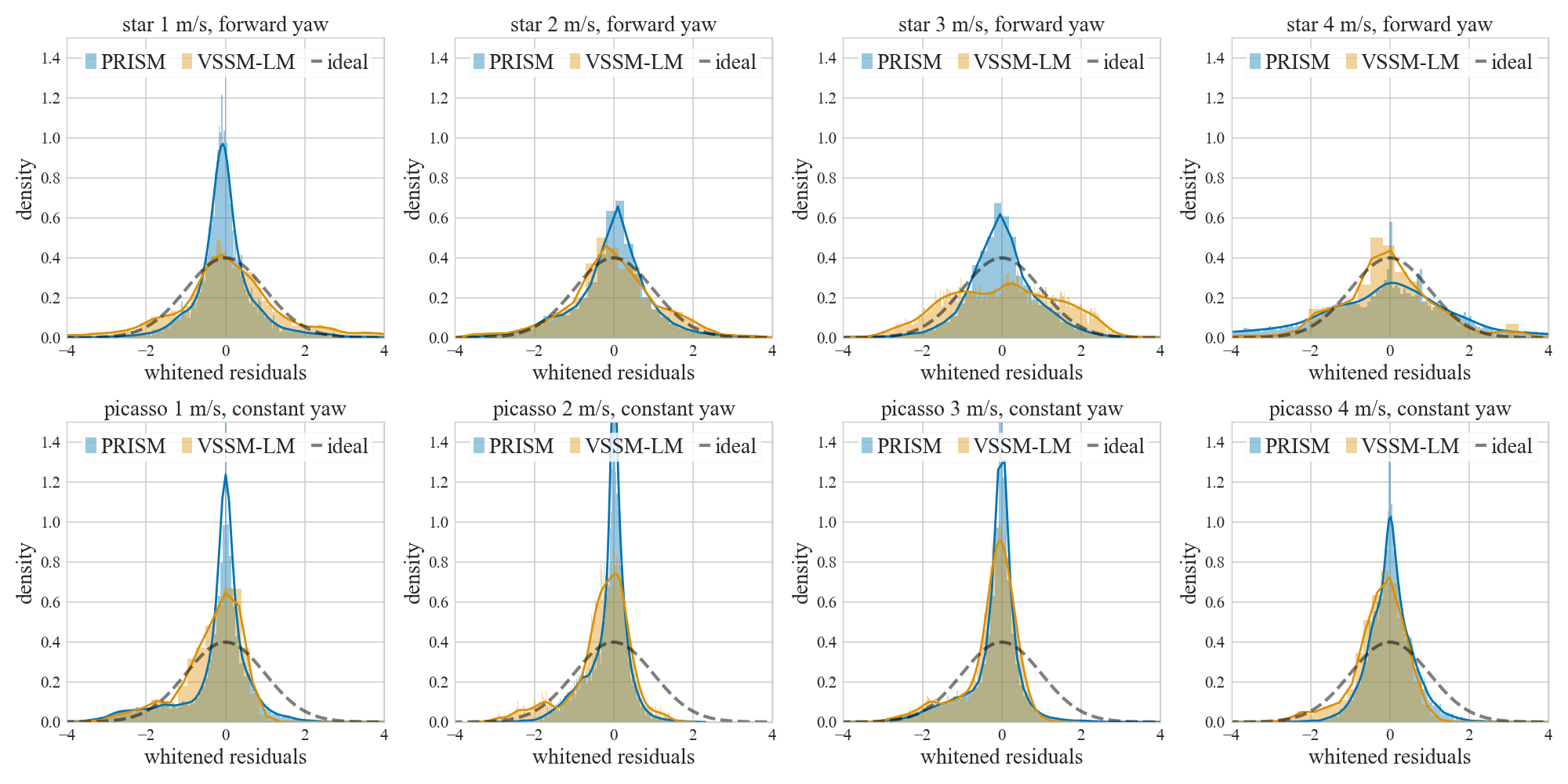}
    \caption{
        Whitened state residuals for PRISM and VSSM-LM \citep{mirchev2021variational} for eight different blackbird trajectories (denoted in the titles).
        The model is pessimistic when the distribution of whitened residuals is narrower than a standard Gaussian.
    }
    \label{fig:whiteresiduals}
\end{figure}
We analyse the uncertainty of PRISM and compare it to our reproduction of the off-line variational inference method of \citet{mirchev2021variational} in JAX, which we denote with VSSM-LM.
We compare both state uncertainty and map uncertainty.
Since we have ground-truth poses for the state, we can also evaluate the state uncertainty calibration quantitatively.
The calibration tells us whether the uncertainty matches the estimation errors between the inferred state means and the ground truth.

\paragraph{State uncertainty}
To evaluate the calibration of state uncertainty, we first evaluate the residuals between the inferred mean poses and the ground-truth MOCAP poses, using the same eight trajectories of the Blackbird data set \citep{blackbird} from \cref{sec:localisation}.
The emission and transition scales of both methods define the overall uncertainty magnitude of the estimates.
To put both systems on equal ground and avoid tuning inefficiencies \wrt these scales, we estimate a single global scalar correction for each method and apply it to all state covariances.
We then whiten\footnote{Normalise by the square root of the covariance matrix, i.e. inverse of a triangular matrix.} the computed residuals with their respective covariance estimates for all poses in all trajectories.

The whitened residuals of a perfectly calibrated model should form a standard normal distribution.
This would indicate that the inferred covariances exactly match the distribution of errors the model makes.
If the whitened residuals end up narrower than a standard Gaussian, then the model is too pessimistic as its uncertainty estimate was higher than the actual unnormalised residuals, and vice versa.
Figure \cref{fig:whiteresiduals} shows the distribution of the whitened residuals for both PRISM and VSSM-LM.
While both models are not perfectly calibrated, their residual distributions are still reasonable, as they roughly match the support of the ideal Gaussian.
PRISM is more pessimistic, indicated by the narrower distributions of whitened residuals.
We find this is better than the alternative, as it would lead to more cautious control of the agent.
Note that PRISM is also consistently more accurate in state estimation than VSSM-LM, as shown in \cref{sec:localisation}.
The pessimism is more pronounced for trajectories of lower velocity (see titles of \cref{fig:whiteresiduals}), for which state estimation is easier.

\begin{wrapfigure}{r}{0.5\linewidth}
    \centering
    \includegraphics[width=\linewidth]{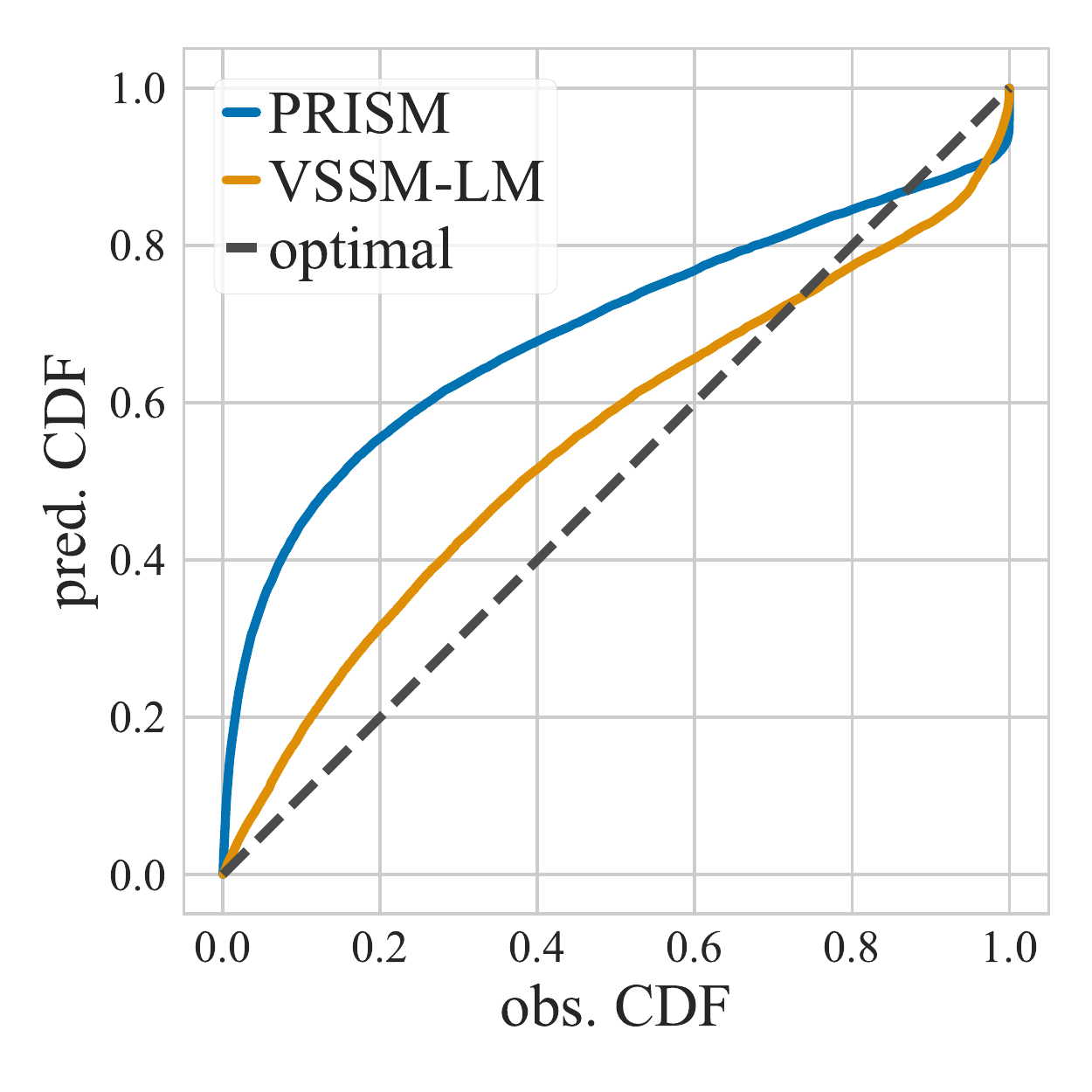}
    \caption{Chi-squared calibration curve.}
    \label{fig:chianalysis}
\end{wrapfigure}

\newpage
To quantify the calibration gap further, we perform a Chi-squared calibration analysis of the normalised squared sum of residuals (NSSR), following \citet{jospin2022hands}, section VII.
Sum here refers to summing over the pose dimension.
\Cref{fig:chianalysis} shows the result.
It compares the cumulative Chi-squared distribution of the normalised residuals (prediction distribution) to the cumulative distribution of observing that residual in the data (observation distribution).
A model with an ideal calibration of its uncertainties relative to the errors it makes would match the identity diagonal.
The x-axis corresponds to an ordering of the magnitude of the residuals (low to high).
Model curves above the diagonal mean the model is pessimistic, and vice versa.
The calibration curves of both methods indicate a reasonable correlation between the cumulative prediction distribution and the cumulative observation distribution of residuals, with PRISM exhibiting more bias towards pessimism.
This is a confirmation of what we identified in the residual plots above.
Note that the uncertainty of VSSM-LM is produced offline, ca. 15 times slower than PRISM in our implementation.

\paragraph{Map uncertainty}
\begin{figure}[t]
    \centering
    \includegraphics[width=\linewidth]{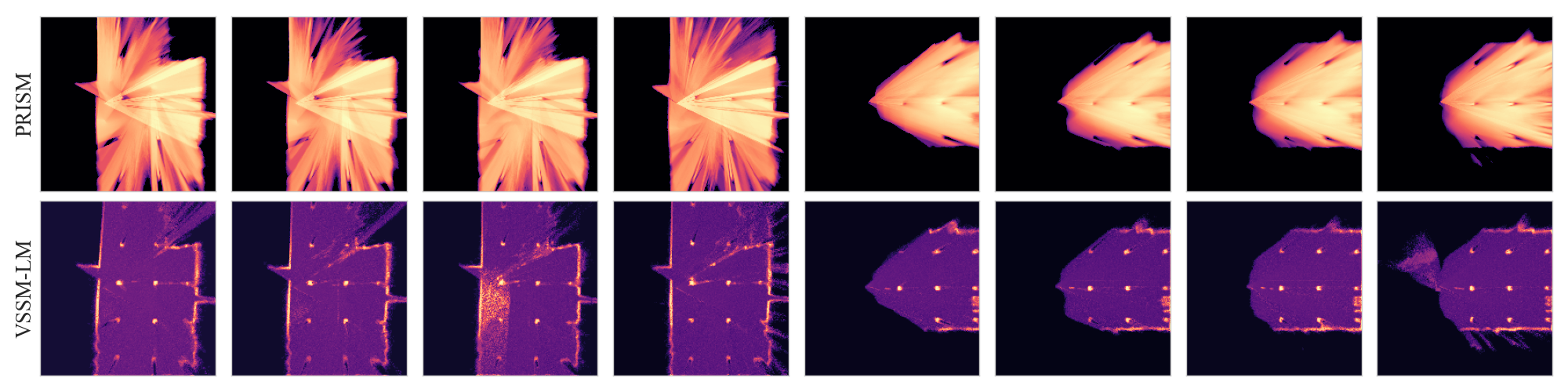}
    \caption{
        Horizontal slices of map uncertainty (orange means certain) for the same eight blackbird trajectories.
        PRISM on top, VSSM-LM below.
        From left to right: \emph{star} \{1, 2, 3, 4\} m/s, forward yaw, then \emph{picasso} \{1, 2, 3, 4\} m/s, constant yaw.
    }
    \label{fig:mapuncerts}
\end{figure}
Next we qualitatively compare the map uncertainty of PRISM to the map uncertainty of VSSM-LM.
\Cref{fig:mapuncerts} shows horizontal slices of the uncertainty estimates at eye level for the same eight Blackbird trajectories from \cref{sec:localisation}.
For all trajectories, the scene is a subway station with multiple columns.

VSSM-LM is certain primarily at surfaces, i.e. at walls and columns (the \color{orange} orange \color{black} lines and dots in the images on the lower row).
It assigns lower certainty to the empty space inside the scene (\color{violet} purple\color{black}), which is technically observed just as much by the agent.
We attribute this to the way differentiable rendering works in this method: since the renderer identifies a single hit along each ray when observations are reconstructed, gradients flow only to the map parameters that correspond to the hit.
Therefore certainty is most pronounced there.

On the other hand, PRISM assigns certainty both to surfaces and to the observed empty space between (\cf \cref{fig:mapuncerts}, top row), due to the nature of the map updates.
Note that whenever something is occluded in the view of the agent it remains less certain (\eg the triangular patterns behind columns seen in the images).
Also, whenever the agent spends more time observing a given region, the certainty of the map increases proportionally.
This can be seen in all star trajectories, where the agent starts its flight from the same spot after sitting still for a while (first four images in \cref{fig:mapuncerts}, also see \cref{fig:mapuncertaintyblackbird}).
Whether this characteristic difference between the methods matters is left to explore in future work.

Overall, both methods reliably leave regions out of view uncertain and increase certainty only in the observed space.
\section{Map Resolution Ablations}
In this ablation we evaluate how localisation accuracy changes for increased map resolution.
We set the new resolution to 400 $\times$ 400 $\times$ 400, doubling the resolution of each voxel grid side.
We fix all other hyperparameters to their default values, as listed in \cref{app:experiments}.
The localisation results are in \cref{table:localisation400}.
We notice consistent improvements across the board, the largest of up to 3 cm for the Blackbird trajectories.
This makes sense, as the Blackbird scenes are the biggest (25 m $\times$ 25 m $\times$ 25 m), where an increased resolution leads to a more substantial reduction in voxel size, from 12.5 cm to 6.25 cm per side.

We notice only one outlier, \emph{star, 4 m/s}.
For the increased resolution, five seeds for that trajectory resulted in RMSE scores of \{0.300, 0.281, 0.420, 0.115, 0.269\} m.
None of the runs failed and they all showed low to moderate estimation bias around the corners of the trajectory.

We note that voxel grids are wasteful in terms of memory and limit the maximum feasible resolution.
Schemes like voxel hashing \citep{niessner2013real} can lift this limitation, and fit all other presented assumptions.

\begin{table}[H]
    \centering
    \footnotesize
    \setlength{\tabcolsep}{2pt}
    \centering
    \captionof{table}{Localisation absolute error RMSE in meters on EuRoC \citep{euroc}, Blackbird \citep{blackbird} and TUM-RGBD~\citep{tumrgbd}, for a map resolution of 400 x 400 x 400 (twice as big as the default).}
    \label{table:localisation400}
    \begin{tabular}{lcc}
        \multirow{2}{*}{Trajectory} & PRISM & PRISM \\
        & res. 200 & res. 400 \\
        \toprule
         EuRoC/V101 &              0.041 ($\pm$ 0.002) &              \textbf{0.038} ($\pm$ 0.003) \\
         EuRoC/V102 &              0.035 ($\pm$ 0.002) &              \textbf{0.030} ($\pm$ 0.001) \\
         EuRoC/V103 &              0.042 ($\pm$ 0.002) &     \textbf{0.038 ($\pm$ 0.002)} \\
         EuRoC/V201 &              0.037 ($\pm$ 0.001) &     \textbf{0.032 ($\pm$ 0.001)} \\
         EuRoC/V202 &     \textbf{0.035 ($\pm$ 0.003)} &     \textbf{0.035 ($\pm$ 0.003)} \\
         EuRoC/V203 &                                x &                                x \\
        \bottomrule
        \addlinespace[1ex]
        \multirow{2}{*}{Trajectory}  & PRISM & PRISM \\
        & res. 200 & res. 400 \\
        \toprule
         picasso, 1 m/s &                   0.064 ($\pm$ 0.003) &                   \textbf{0.045 ($\pm$ 0.003)} \\
         picasso, 2 m/s &                   0.053 ($\pm$ 0.003) &                   \textbf{0.048 ($\pm$ 0.009)} \\
         picasso, 3 m/s &                   0.061 ($\pm$ 0.003) &                   \textbf{0.042 ($\pm$ 0.002)} \\
         picasso, 4 m/s &               0.079 ($\pm$ 0.005)\tablefootnote{\label{blackbirdnote2}Last  10 s are skipped, as the drone hits the ground during landing.} &     \textbf{0.061 ($\pm$ 0.002)}\footref{blackbirdnote2} \\
         star, 1 m/s    &      0.089 ($\pm$ 0.007)\footref{blackbirdnote2} &      \textbf{0.074 ($\pm$ 0.009)}\footref{blackbirdnote2} \\
         star, 2 m/s    &                   0.111 ($\pm$ 0.009) &                   \textbf{0.103 ($\pm$ 0.009)} \\
         star, 3 m/s    &          0.115 ($\pm$ 0.012) &          \textbf{0.082 ($\pm$ 0.008)} \\
         star, 4 m/s    &          \textbf{0.153 ($\pm$ 0.015)}\footref{blackbirdnote2} &          0.278 ($\pm$ 0.015) \footref{blackbirdnote2} \\
        \bottomrule
        \addlinespace[1ex]
        \multirow{2}{*}{Trajectory} & PRISM & PRISM \\
        & res. 200 & res. 400 \\
        \toprule
        fr1/desk   &         0.053 ($\pm$ 0.003) &         \textbf{0.052 ($\pm$ 0.001)} \\
        fr2/xyz    &         0.029 ($\pm$ 0.001) &         \textbf{0.021 ($\pm$ 0.000)} \\
        fr3/office &          0.083 ($\pm$ 0.001) &          \textbf{0.081 ($\pm$ 0.002)} \\
        \bottomrule
    \end{tabular}
\end{table}
\section{Motivation and Downstream Applicability} \label{app:motivation}

PRISM is an inference tailored to the state-space model introduced by \citet{mirchev2021variational}, and this synergy has advantages.
Markovian state-space models are a fundamental building block that enables model-based control \citep{bertsekas}.
For this, both a predictive distribution and a state estimator are needed.
The predictive distribution is already given by \citet{mirchev2021variational}, PRISM fills the role of a real-time state estimator.

The advantages we foresee come from the probabilistic integration of a dense map, rendering and dynamics.

\paragraph{Prediction and Control}
PRISM's estimates harmonise with the predictive model:
\eq{
    \mathbb{E}_{\qfilter{t}{\map}\qfilter{t}{\state_t}}[\pp{\state_{t+1:t+k}, \obs_{t:t+k}}{\control_{t:t+k-1}, \state_t, \map}],
}
because we derive them as approximations to posterior distributions stemming from the same state-space model.
A rollout can therefore start from the inference estimates $\qfilter{t}{\map}$ and $\qfilter{t}{\state_t}$ (in expectation above) and predict both rendered images and states with appropriate uncertainty for a candidate control sequence~$\control_{t:t+k-1}$.
We can then apply any technique from the literature on optimal control and reinforcement learning to control the agent \citep{bertsekas,sutton2018reinforcement}.
Note that because the control problem is of partial observability, \ie a partially-observable Markov decision process (POMDP), an ideal solution may need to intertwine the estimator in the rollout to form beliefs, but we leave such considerations for the future to simplify discussion.

One advantage we see is that whole images can be predicted in the rollout.
This is possible because of the dense map estimate, which supports rendering.
It allows us to define rewards for the image observations, which can enable interesting control tasks that are not limited to point-to-goal navigation.

Another advantage is that the dense map directly provides obstacle information.
We can combine this information with the map uncertainty estimates to be more robust when avoiding obstacles in the environment. Similarly, since the predictive rollout is fully-probabilistic, the uncertainty of $\qfilter{t}{\state_t}$ will propagate through the state transition and would be useful for robust collision avoidance as well.

On its own, the uncertainty of the map is useful for active learning (\eg see \citep{abise}).
In particular, we can go beyond frontier-based exploration and focus the agent on still uncertain map regions through an information-theoretic objective.

\paragraph{Adoption and Future Work}
We have made an effort to signpost all approximations we make in our filtering derivations in \cref{app:mapfilter,app:statefilter}.
We hope this will facilitate future research.

PRISM is also straightforward to implement in auto-differentiable frameworks (the current version is written entirely in JAX \citep{jax2018github}).
  
\end{document}